\def\year{2021}\relax
\documentclass[letterpaper]{article} 
\usepackage{aaai21}  
\usepackage{times}  
\usepackage{helvet} 
\usepackage{courier}  
\usepackage[hyphens]{url}  
\usepackage{graphicx} 
\usepackage{natbib}  
\usepackage{amsmath,amssymb,amsfonts,amsthm}

\usepackage{diagbox}
\usepackage{nameref}

\usepackage{hyperref}
\usepackage{color,xcolor}
\usepackage[ruled,vlined,linesnumbered]{algorithm2e}
\usepackage{algorithm2e}
\usepackage{algorithmic}
\usepackage{multirow}
\usepackage{helvet} 
\usepackage{courier}  
\usepackage[caption=false]{subfig}
\DeclareMathOperator*{\argmin}{arg\,min}
\DeclareMathOperator*{\argmax}{arg\,max}
\newtheorem{theorem}{Theorem}
\newtheorem{lemma}{Lemma}
\newtheorem{remark}{Remark}
\newtheorem{observation}{Observation}[theorem]
\newtheorem{definition}{Definition}
\newtheorem{corollary}{Corollary}[theorem]
\usepackage{xspace}

\begin{document}
	\title{Characterizing the Evasion Attackability of Multi-label Classifiers}
	\author{Zhuo Yang$^1$, Yufei Han$^2$, Xiangliang Zhang$^1$\\}
	\affiliations{
		$^1$King Abdullah University of Science and Technology, Thuwal, Saudi Arabia\\ 
		$^2$Norton Research Group, Sophia Antipolis, France\\ 
		$\{ $zhuo.yang, xiangliang.zhang$\}$@kaust.edu.sa\\
		yfhan.hust@gmail.com\\
	}
	\maketitle
	
	\begin{abstract}
		Evasion attack in multi-label learning systems is an interesting, widely witnessed, yet rarely explored research topic. Characterizing the crucial factors determining the attackability of the multi-label adversarial threat is the key to interpret the origin of the adversarial vulnerability and to understand how to mitigate it. Our study is inspired by the theory of adversarial risk bound. We associate the attackability of a targeted multi-label classifier with the regularity of the classifier and the training data distribution. Beyond the theoretical attackability analysis, we further propose an efficient empirical attackability estimator via greedy label space exploration. It provides provably computational efficiency and approximation accuracy. Substantial experimental results on real-world datasets validate the unveiled attackability factors and the effectiveness of the proposed empirical attackability indicator. 
	\end{abstract}
	
	\section{Introduction}\label{sec:introduction}
	
	Evasion attack has been witnessed widely in real-world practices of \emph{multi-label learning} \cite{SongQi2018ICDM}. For example, a creepware/stalkware usually has multiple malicious labels as it sniffs the victim's privacy via different mobile services. To avoid being flagged, the entities authoring these malwares \cite{kevin2020sp,Ristenpart2018chi} tend to hide their key malicious labels, such as rending remotely recording phone calls or accessing private files, by slightly reprogramming app binary structures. Meanwhile, they preserve less harmful labels like GPS tracking to pretend to be benign parental control. Another example can be found in image recommendation systems. An adversary tends to embed spam/toxic advertisements \cite{Gupta2013www} into a recommended image with other harmless contents. These malicious contents are so well tuned that the sanitary check system is deceived by the camouflaged image, while recognizing correctly other harmless scenarios. 
	
	Despite of the widely existence of multi-label adversarial threats, it has been a rarely investigated, yet important topic to evaluate the robustness of a multi-label classifier under evasion attack (a.k.a. \textbf{attackability}).
	Intuitively, assessing the attackability of a multi-label classifier $h$ with an input instance is to explore the maximal perturbation on $h$'s output that an input adversarial noise of bounded magnitudes can ever cause. The problem of attackability assessment in a general setting can be  defined as: \textbf{given a magnitude bound of the adversarial perturbation and the distribution of legal input data instances,  what is 
		the worst-case miss-classification risk of $h$ under the attack?} 
	Classifier $h$ is more attackable if it has a higher risk, while $h$ is certified to be not attackable  if its output cannot be changed by any adversarial noise within the magnitude bound.
	Via attackability assessment, we aim at answer the following questions:
	\begin{itemize}
		\item  {What are the factors determining the attackability level of a multi-label classifier?} 
		\item  {Can we derive an empirically computable attackability measurement for a multi-label classifier?}  
	\end{itemize}
	
	Echoing the questions raises two challenges:  {First}, analyzing the worst-case classification risk on adversarial instances with PAC-learning framework requires a fixed distribution of adversarial instances. However, it is a well received fact that such a distribution depends radically on the targeted classifier's property, thus it is not fixed and closely associated with the classifier's architecture. The celebrated works \cite{Yin2018icml,JKhim2018ft,TuNIPS19} proposed to mitigate the gap via the lens of Rademacher complexity. Nevertheless, they all focused on single-label  scenarios, thus can't be applied  to answer the questions above.  {Second}, evaluating the worst-case risk for an input   instance needs to explore the maximal set of jointly attackable labels. Since labels are not mutually exclusive in multi-
	label tasks, the label exploration process is in nature an NP-hard mixed-integer programming problem. The adversarial noise generation method in \cite{SongQi2018ICDM} applies only in the targeted attack scenario, where the attacked labels are given. Few effort has been dedicated to study the feasibility of label space exploration.
	

	To address the challenges above, we conduct both theoretical and empirical attackability analysis of a multi-label classifier. Our contributions can be summarized as follows:
	\begin{itemize}
		\item We measure the attackability of a multi-label classifier by evaluating the expected worst-case miss-classification loss over the distribution of adversarial examples. We instantiate the study to linear and deep neural networks, 
		which are popularly deployed in multi-label applications. Our analysis unveils that the attack strength, the regularity of the targeted classifier and the empirical loss on unperturbed data are the external and intrinsic driving force jointly determining the attackability level. We further  reveal the theoretical rationality of the low-rank regularization and adversarial training in hardening the classifier. 
		

		
		\item 
		We cast the problem of evaluating the empirical attackability level as \emph{a label space exploration process} for each of the legal input instances. We further demonstrate the triviality of the label space exploration problem by formulating it as a submodular set function optimization problem. Thus it can be solved via greedy search with certified approximation accuracy.

		\item  We propose a \emph{Greedy Attack Space Expansion} (GASE) algorithm to address the computational bottleneck of the primitive greedy search for empirical attackability measurement. The proposed method provides a computationally economic estimator of the marginal gain obtained by adding a candidate label into the  set of attacked labels. It selects the labels with the largest marginal gain to deliver an efficient exploration of attack targets. 
		
		
		
	\end{itemize}

	\section{Related works}
	
	\textbf{Robustness against adversaries.} The emergence of evasion attack raises a severe challenge to practitioners’ trust on machine learning in performance-critic applications \cite{biggio2018pr,BiggioECML2013,CarliniSP2017}. Considerable efforts have been dedicated to detect adversarial samples, improve model designs and propose robust training methods \cite{Cullina2018nips,Goodfellow2015,Athalye2018icml,AFawzi2016nips,Szegedy2013IntriguingPO,Xu2018ndss,Madry2018iclr,Ross2018ImprovingTA,Jakubovitz2018eccv,Matthias2017nips,Zugner2019CertifiableRA,Bojchevski2019nips,Raghunathan2018nips,Tramr2018EnsembleAT,Papernot2016DistillationAA,Zugner2020kdd,cohen2019icml,Lee2019nips}. Especially, \cite{wang2019pmlr,Shafahi2019nips,Gao2019nips} discussed convergence guarantee and high sample complexity of adversarial training. In contrast, few literature focuses on the essential problem of \emph{evaluating the vulnerability of a classifier under a given evasion attack setting and identifying key factors determining the feasibility of evasion attack against the targeted classifier}. Pioneering works of this topic  \cite{Matthias2017nips,Wang2018AnalyzingTR,AFawzi2016nips,Gilmer2018arxiv} focused on identifying the upper bound of adversarial noise, which guarantees the stability of the targeted classifier's output, a.k.a adversarial sphere. Notably, \cite{AFawzi2016nips} pointed out  the association between adversarial robustness and the curvature of the classification boundary. Strengthened further by \cite{Yin2018icml,JKhim2018ft,TuNIPS19}, the expected classification risk under adversarial perturbation can be bounded by the Rademacher complexity of the targeted classifier. Moreover in \cite{QiSysML2018,Han2020kdd}, attackability of a recurrent neural net based classifier on discrete inputs   was measured by checking the regularity of the targeted classifier. 
	Apparently, the regularity of the targeted classifier play an equally important role as the attack strength in causing adversarial vulnerability. Inspiring as they are, these works focus on single-label learning tasks. 
	Due to the label co-occurrence in multi-label learning, searching for the combinations of attacked labels causing the worst-case loss is NP-hard. It is thus an open issue to evaluate the adversarial risk of multi-label learners. 
	
	
	\noindent
	\textbf{Noise-tolerant multi-label learning.}\label{sec:related_work}
	A relevant topic is to learn multi-label classifier with imperfect training instances. Miss-observations and noise corruptions of features and labels of training instances can introduce severe bias into the derived classifier. Most research efforts in this domain recognised that   the key to success is to encode label correlation and the predicative relation between features and labels
	\cite{YuYinSun:2010,Zhu:2010,Liu:2010,Lin:2013,Wu:2013,Zhao:2015,Yu:2014,BiW:2014,Goldberg:2010,Cabral:2015,XuMiao:2013,Chiang:2015,Guo17:2017,YueZhu:2018,Hsieh:2015}. 
	They exploited not only low-rank structures of feature/label matrices 
	for missing data imputation, but also gained stable performances by enforcing the low-rank regularization on the predictive model capturing the feature-label correlation. 
	Especially \cite{Xu2016TIP} proposed to regularize the local Rademacher complexity of a linear multi-label classifier in the training process. The study indicated the link between the Rademacher complexity and the low-rank structure of the classifier's coefficients. The reported results showed that a lower-rank structured linear classifier can better recover missing labels. Nevertheless, all the previous works focus on adversary-free scenarios. Furthermore, the analysis over the role of low-rank structures was limited to linear multi-label classifiers. It is thus interesting to study whether the low rank driven regularization can help to mitigate the adversarial threat against both linear and DNN based multi-label classifiers.
	
	


	\section{ Attackability of Multi-label Classifiers}\label{sec:methods}
	\subsubsection{Notations and Problem Definition.}
	We assume $\mathcal{Z} = \mathcal{X} \times \mathcal{Y}$ as a measurable multi-label instance space, with $\mathcal{X} \in {\mathbb{R}^d}$ and $\mathcal{Y} = {\{ - 1,1\} ^m}$, where $d$ is the feature  dimension  and $m$ is the number of labels. 
	Given $n$ i.i.d. training examples $ \{ ({\bf{x}}_i,{\bf{y}}_i)\} $  drawn  from $\mathcal{P}(\mathcal{Z})$, the classifier $h \in \mathcal{H}:\mathcal{X} \to \mathcal{Y}$ is learnt by minimizing the empirical loss  $\sum_{i=1}^{n}\ell({\bf{x}}_i,{\bf{y}}_i)$, with the loss function $\ell:\mathcal{X} \times \mathcal{Y} \to {\mathbb{R}}$. 
	
	Eq.(\ref{eq:mlevasion}) defines a typical scenario of  empirical attackability evaluation for a multi-label classifier $h$ given an input $\bf{x}_i$,  perturbed by $\bf{r}$.  
	The 
	classification output  $sgn(h({\bf{x}}_i+\bf{r^{*}}))$ has been   flipped on as many as possible labels. 
	\begin{equation}\label{eq:mlevasion}
	\small
	\begin{split}
	&C^{*}({\bf{x}}_{i}) = \underset{T,\|{\bf r}^{*}\|\leq{\mu_{r}}}{\mathbf{max}}\,\,\,\sum_{j=1}^{m}I({ y}_{ij}{\neq}sgn(h_{j}({\bf x}_i+{\bf r}^{*}))),\\
	&{\text{where}} \,\,\,{\bf r}^{*} = \argmin_{\bf r}\|{\bf r}\|,\\
	\,&s.t.  \,\,y_{ij}h_{j}({\bf x}_i+ {\bf r}^* ) \leq 0\,\,(j\in{T}),\, \,\,y_{ij}h_{j}({\bf x}_i+{\bf r}^*) > 0\,  (j\notin{T}).\\
	\end{split}    
	\end{equation}
	where $h_{j}({\bf x}_i+\bf{r})$ denotes the classification score for the label $j$ of the adversarial example, and $sgn$ is the sign function outputting $\pm{1}$ based on the sign of $h_{j}({\bf x}_i+\bf{r})$. The indicator function $I(\cdot)$  outputs 1 if the attack flips a label and 0 otherwise. $T$ denotes the set of flipped labels. The magnitude of $C^{*}$ indicates the attackability of the classifier given the attack strength limit $\mu_r$ and the input ${\bf{x}}_i$.  Given the same input $\bf{x}$ and the bound of  perturbation  $\mu_r$, one multi-label classifier $h$ is more attackable than the other $h'$, if $C^{*}_{h}>C^{*}_{h'}$. 
	
	

	
	\subsection{Bound of Adversarial Attackability}
	Beyond the attackability measurement given a local fixed input instance, we pursue  {a theoretical and empirical insight into the attackability of $h$ in the space of adversarial samples}, which are sampled 
	from a new distribution $\mathcal{P}'$ translated from $\mathcal{P}$ after injecting the adversarial perturbation. 
	The distribution shift from $\mathcal{P}$ to $\mathcal{P}'$ is the origin of adversarial threat, as it violates the i.i.d. assumption of the learning process. 
	By assuming that $\mathcal{P}'$ lies within a Wassernstein ball centered at $\mathcal{P}$  with a radius of $\epsilon$, we have the following definition about classification risk under   evasion attack. 
	\begin{definition}\label{emprisk}
		For a multi-label classifier $h$ and legal input samples $\{{\bf x}_i,{\bf y}_i\}\sim{\mathcal{P}}$ and its
		corresponding adversarial samples 
		$\{{\bf x}'_i,{\bf y}_i\}\sim{\mathcal{P}'}$, 
		{the worst case expected and empirical risk under the evasion attack} are: 
		\begin{equation}\label{emrisk}
		\small
		\begin{split}
		{R_{\mathcal{P'}}}(h) &= {E_{(\bf{x},\bf{y})\sim{\mathcal{P}}}} {[\mathop {\max }\limits_{{(\bf{x}',\bf{y})\sim{\mathcal{P}'}},\mathcal{W}(\mathcal{P},\mathcal{P}')\leq{\epsilon}} l(h({{\bf{x}'}}),{\bf{y}})]} ,\\
		{R^{emp}_{\mathcal{P'}}}(h) &= \frac{1}{n}\sum\limits_{i = 1}^n {[\mathop {\max }\limits_{{({\bf{x}'}_{i},{\bf{y}}_{i})\sim{\mathcal{P}'}},\mathcal{W}(\mathcal{P},\mathcal{P}')\leq{\epsilon}} l(h({{\bf{x}'}_i}),{{\bf{y}}_i})]}, \\
		\end{split}
		\end{equation}
		where $\mathcal{W}(\mathcal{P}',\mathcal{P})$ denotes the Wassernstein distance between $\mathcal{P}'$ and $\mathcal{P}$, and 
		$\epsilon$ is  the radius of the adversarial space. 
	\end{definition}
	The $\mathcal{W}(\mathcal{P}',\mathcal{P})$
	can be bounded with the magnitude of the adversarial perturbation after \cite{TuNIPS19}, which gives $\mathcal{W}(\mathcal{P}',\mathcal{P})\leq \sup_{{\bf{x}'},{\bf{x}}}\|{{\bf{x}'}_i}-{{\bf{x}}_i}\|_{2}\leq{\mu_r}$.  {Without loss of generality, we use the Euclidean distance $\|{{\bf{x}'}_i}-{{\bf{x}}_i}\|_{2}$ to constrain the attack budget hereafter}. Consistent with the defined  attackability evaluation scenario in Eq.\ref{eq:mlevasion}, $R_{\mathcal{P}'}(h)$ 
	measures the attackability of $h$. A higher $R_{\mathcal{P}'}(h)$ indicates a more attackable $h$. And 
	${R^{emp}_{\mathcal{P'}}}(h)$ is the empirical estimator of the attackability level. By definition, if we derive $C^{*}$ by solving Eq.(\ref{eq:mlevasion}) for each instance $({\bf{x}}_{i},{\bf{y}}_{i})$ and adopt the binary 0-1 loss, an aggregation of the local worst-case loss  $\sum_{i=1}^{n} C^{*}({\bf x}_{i})$ gives ${R^{emp}_{\mathcal{P'}}}(h)$. 
	
	In the followings, we establish the upper bound of the attackability measurement with respect to \emph{ linear and feedforwad neural network multi-label classifiers}. It reveals the key factors determining the attackability level of a classifier. 
	Given ${\bf x} \in {\mathbb{R}^d}$ and ${\bf{y}}\in   {\{ - 1,1\} ^m}$ as the feature and label vector of a data instance, a linear multi-label classifier is $h(\bf{x}) = \bf{x}w$. The linear coefficient matrix  ${\bf w}\in{\mathbb{R}^{d\times{m}}}$ is defined with the spectral norm   $\|{\bf w}\|_{\delta}\leq{\Lambda}$. Furthermore, we constrain  the range of legal inputs and the adversary's strength as $\|{\bf{x}}\|_{2}\leq{\mu_{x}}$ and $\|{\bf r}\|_2\leq{\mu_{r}}$ respectively. Without loss of generality, a Least-Squared Error (LSE) loss is adopted to compute the classification risk, such as  $\ell({\bf  x , y ) = \| y  -   x w}\|_{2}$. 
	A distance metric for $z  = \{{\bf x} ,{\bf y} \}$ is defined as $d(z_i,z_j) = \|{\bf x}_i -{\bf x}_j\|_{2} + \|{\bf y}_i - {\bf y}_j\|_{2}$. 
	
	\begin{theorem}\label{theorem:linear}
		\textbf{[Upper-bound of attackability for a linear multi-label classifier]} 
		The upper bound of $R_{\mathcal{P}'}(h)$ 
		holds with at least probability of 1-$\sigma$: 
		\begin{equation}\label{eq:upperboundlinear}
		\small
		\begin{split}
		&R_{\mathcal{P}'}(h) \leq R^{emp}_{\mathcal{P}'}(h) +
		96\sqrt{\frac{\mu_{x}{\Lambda}R(1+\mu_{x}\Lambda)}{n}} \\
		&+ \frac{12{C_{h}}\sqrt{\pi}(m+2\mu_{x})}{\sqrt{n}} + (m+\Lambda{\mu_{x}})\sqrt{\frac{\log(1/\sigma)}{2n}}, 
		\end{split}
		\end{equation}
		and the worst case empirical loss has the upper bound: 
		\begin{equation}\label{eq:upperboundlinear1}
		\centering
		\small
		R^{emp}_{\mathcal{P}'}(h) \leq \frac{1}{n}\sum_{i=1}^{n}\ell(h({\bf x}_i),{\bf y}_i) + C_{h}\mu_{r},
		\end{equation}
		where $R$ denotes the rank of the coefficient matrix $\bf w$ and $C_{h} = {max\{\|{\bf w}\|_{\delta},1\}}$.  
		
	\end{theorem}
	The proof is presented in supplementary document.
	
	\begin{remark}\label{remark1}
		We have three observations from the derived analysis in Eq.(\ref{eq:upperboundlinear}-\ref{eq:upperboundlinear1}). \\
		1) The  worst-case empirical loss $R^{emp}_{\mathcal{P}'}$ can be used as a sensitive indicator of the worst-case expected adversarial risk $R_{\mathcal{P}'}$. Thus it can be used as an empirical measure of attackability of the classifier over the adversarial data space. A lower $R^{emp}_{\mathcal{P}'}$ 
		implies a lower expected miss-classification risk in the adversarial space. \\
		2) The spectrum of linear coefficient matrix $\bf w$ plays an important role in deciding the attackability level of $h$. 
		Especially, 
		$h$ with lower rank $\bf w$ has lower expected miss-classification risk. This is consistent with what was unveiled in previous research of multi-label classification: enforcing low-rank constraints over the linear    classifier usually brings robustness improvement against noise corruption. \\
		3) \emph{The empirical risk over unperturbed data}, \emph{the magnitude of the adversarial perturbation} and \emph{the spectrum of the classifier's coefficient matrix} are the three main factors jointly determining the attackability of the multi-label classifier. {On one hand}, the risk upper bound depends on the external   driving force of the adversarial threat, which is the   magnitude of the adversarial perturbation $\|\mu_{r}\|$.
		{On the other hand}, the internal factors on the riks upper bound are the regularity of the classifier (low-rank structure) and  the profile of the training data distribution. 
		Moreover, by dropping the terms with $\mu_{r}$ in Eq.(\ref{eq:upperboundlinear}), we can find that an adversary-free generalization bound of the linear multi-label classifier heavily depends on the low rank structure of the classifier. It is consistent with the results unveiled by previous works \cite{XuMiao:2013,Yu:2014,YueZhu:2018}: low-rank structured classifiers are favorable in multi-label classification. Due to the page limit, we leave this discussion in the supplementary document.  
		
	\end{remark}

	Inherited the setting of the attack scenario from Theorem.\ref{theorem:linear}, we consider a neural network based multi-label classifier $h_{nn}$ with $L$ layers, where:
	\begin{itemize}
		\item The dimension of each layer is $d_{1},d_{2},...,d_{L}$, and $d_{0} = d$ for taking input $\bf x$ 
		and $d_{L} = m$ for outputting labels $\bf y$ .  
		\item At each layer $i$, $A_{i}\in{R^{d_{i-1}{\times}d_{i}}}$ denotes the linear coefficient matrix (connecting weights). The spectral norm of $A_{i}$ is bounded as $\|A_{i}\|_{\delta}\leq{\Lambda_{i}}$. $R_{i}$ denotes the rank of $A_{i}$. 
		\item The activation functions used in the same layer are Lipschitz continuous and share the same Lipschitz constant $\rho_{i}$. We use $g_{i}$ to denote the activation functions used at the layer $i$. The output of each layer $i$ can be defined recursively as $\mathcal{H}_{i} = g_{i}(\mathcal{H}_{i-1}A_{i})$.
	\end{itemize}
	\begin{theorem}\label{theorem:fnn}
		\textbf{[Upper-bound of attackability for neural nets based multi-label classifier]} 
		The upper bound of ${R_{\mathcal{P}'}}(h_{nn})$ holds with at least probability of $1-\sigma$: 
		\begin{equation}\label{eq:upperboundnn}
		\small
		\begin{split}
		&R_{\mathcal{P}'}(h_{nn})\leq R^{emp}_{\mathcal{P}'}(h_{nn}) + 2m\sqrt{\frac{\log(1/\sigma)}{2n}} + \\
		&\frac{96\sqrt{dm\Lambda_d}\sum_{i=1}^{L}{R_i}\sqrt{d_{i}{\Lambda_i}C_i}}{\sqrt{n}} + \frac{12{C_{nn}}(2\mu_{x}+m)\sqrt{\pi}}{\sqrt{n}}   
		\end{split},
		\end{equation}
		and the empirical loss $R^{emp}_{\mathcal{P}'}$ has the upper bound: 
		\begin{equation}\label{eq:upperboundnn1}
		\small
		\centering
		R^{emp}_{\mathcal{P}'}(h_{nn}) \leq \frac{1}{n}\sum_{i=1}^{n}\ell(h_{nn}({\bf x}_i),{\bf y}_i) + C_{nn}\mu_{r},
		\end{equation}
		where $C_{1} = \rho_{d}$ and $C_{i} = \prod_{j=d-i+1}^{d}\rho_{j}\prod_{j=d+2-i}^{d}\|A_{j}\|_{\delta}$ with $i\geq{2}$, and $C_{nn} = max\{1,\prod_{i=1}^{L}\rho_{i}\|A_i\|_{\delta}\}$.
	\end{theorem}
	The proof is presented in supplementary document.
	
	\begin{remark}\label{remark3}
		Similar to the observations in Remark \ref{remark1}, the attackability of $h_{nn}$ depends heavily on the spectrum of linear coefficients at each layer of the neural nets,  the empirical loss of $h_{nn}$ on legal input samples and the attack strength $\|\mu_{r}\|$. More specifically, the linear coefficient matrices $\{A_{i}\}$ with lower ranks  and lower spectral norm can make $h_{nn}$ more robust. 
		Indeed, enforcing regularization on the spectral norm of the linear coefficients can improve the generalization capability of DNN \cite{Yoshida2017SpectralNR,Miyato2018}. Our analysis not only provides the theoretical rationality behind the reported empirical observations, but also, unveils the impact of the low-rank constraint on $\{A_{i}\}$ in controlling $h_{nn}$'s attackability. 
	\end{remark}
	
	
	
	From Remark 1 and 2, 
	we find that {both reducing the worst-case empirical  risk of the targeted classifier  and enforcing low-rank constraints on its coefficients can help to reduce the attackability and mitigate the risk on evasion attack}. The former can be achieved by conducting adversarial training with the crafted worst-case multi-label adversarial samples. 
	The latter aims at controlling the Rademacher complexity of the targeted classifier, which improves the generalization capability of the targeted classifier. In \cite{XuM2010colt}, the close association between generalization and robustness was explored. Better generalization capability indicates more robustness against noise corruption. 

	\section{Empirical Attackability Evaluation by Greedy Exploration}\label{sec:greedy_method}
	\subsection{Problem Reformulation}
	Solving Eq.(\ref{eq:mlevasion}) to compute the worst-case loss $R^{emp}_{\mathcal{P}'}(h)$ over legal input instances is an NP-hard mixed-integer non-linear constraint problem (MINLP). Traditional solutions to this problem, such as Branch-and-Bound, has an exponential complexity in the worst case. 
	To achieve an efficient evaluation, we propose to empirically approximate $R^{emp}_{\mathcal{P}'}$ via greedy forward expansion of the set of the attacked labels. We re-formulate the label exploration problem in Eq.(\ref{eq:mlevasion}) as a bi-level set function optimization problem:
	\begin{equation}\label{eq:bilevelopt}
	\small
	\begin{split}
	&S^{*} = \underset{S}{\argmax}\,{\psi(S)},\\
	&\text{where}\,\,\,\psi({S}) = \max_{S,|S|\geq{k}}\,\{|S| - g(S)\},\\
	&g(S) = \min_{T\subseteq{S},\|{\bf r}\|\leq{\mu_{r}}}\|{\bf r}\|^{2}_2,\\
	s.t.&\,\,(1-2b_{j})y_{j}h_{j}({\bf x+r})\geq{t_j},\,\,\,j=1,2,...,m,\\
	&\,\,\,b_{j}=1\,\,\,(\text{for}\,\,\,j\in{T}), \,\,\,b_{j}=0\,\,\,(\text{for}\,\,\,j\notin{T}),\\
	\end{split}
	\end{equation}
	where 
	label $y_{j}=\{+1,-1\}$, and $ t_{j}  $ is the minimum classification margin value  enforced on label $j$. The core components of the constraints are the binary indicators $\{b_{j}\}$. With $b_{j}=1$, label $y_j$ is flipped, while with $b_{j}=0$, the label remains unchanged.
	The set function $g(S)$ returns the minimal magnitude of the perturbation $\bf r$ ever achieved via attacking the labels indicated by subsets of $S$.
	In this sense, the inner layer of Eq.(\ref{eq:bilevelopt}) defines an evasion attack against the multi-label classifier targeting at the labels indicated by ${S}$. The optimization objective of the outer layer aims at expanding the set $S$ as much as possible while minimizing as much as possible the required attack cost $\|{\bf r}\|_{2}$. Notably, we set a lower bound $k$ for $|S|$ in Eq.(\ref{eq:bilevelopt}) for the convenience of presentation.
	In a naive way, we can gradually increase the lower bound $k$ until the attack cost valued by $\psi(S)$ surpasses the budget limit. The volume $|S|$ of the derived set gives an estimate of $R^{emp}_{\mathcal{P}'}$ with the binary loss. \\
	\begin{lemma}\label{lemma:submodularity}
		The outer layer of Eq.(\ref{eq:bilevelopt}) defines a problem of non-monotone submodular function maximization. Let $\psi({\hat{S}})$ and $\psi({S^{*}})$ denote respectively the objective function value obtained by randomized greedy forward search proposed in \cite{Buchbinder2014} and the underlying global optimum following the cardinality lower bound constraint. The greedy search based solution has the following certified approximation accuracy:
		\begin{equation}\label{eq:submodular_bound}
		\psi(\hat{S}) \geq \frac{1}{4}\psi(S^{*}).
		\end{equation}
	\end{lemma}

	\subsection{Fast Greedy Attack Space Exploration}
	According to Lemma.\ref{lemma:submodularity}, the set $S$ derived from the random greedy search produces an attack cost $\|{\bf r}\|_{2}$ that is close to the one achieved by the global optimum solution. It guarantees the quality of the greedy search based solution. 
	The \emph{primitive greedy forward expansion} is thus designed as follows:
	\begin{itemize}
		\item We initialize an empty ${S}$ (flipped labels), which assumes no labels are attacked at the beginning. 
		\item In each round of the greedy expansion, for the current set ${S}$ and current adversarial noise ${\bf r}(S)$, we choose each of the candidate labels $j$ out of ${S}$ and compute the marginal gain $\|{\bf r}(S\cup{j})\|_{2} - \|{\bf r}(S)\|_{2}$ by conducting targeted multi-label evasion attack. $\|{\bf r}(S\cup{j})\|_{2}$ is the magnitude of the adversarial noise to flip all the labels in $S\cup{j}$. We then select randomly one of the candidate labels $j$ with the least marginal gains to update ${S} = {S}\cup{j}$.
		\item We update $\|{\bf r}\|_{2}$ by conducting an evasion attack targeting at the labels indicated by ${S}$. The expansion stops when $\|{\bf r}\|_{2}\geq{\mu_{r}}$. 
	\end{itemize}
	
	In each iteration, the \emph{primitive greedy forward expansion} needs to perform evasion attack for each candidate label. It requires \emph{$(m+1)k-k(k-1)/2-1$} evasion attacks before including $k$ labels in $S$. It is costly when the label dimension is high. To break the bottleneck, we propose a computationally economic estimator to the magnitude of the marginal gain $\Delta = \|r({{S}}\cup{j})\|_{2} - \|r({S})\|_{2}$. 
	\begin{lemma}
		In each iteration of the greedy forward expansion, the magnitude of the marginal gain $\Delta$ is proportional to $\frac{\left| y_{j}{{h_j}({\bf{x}} + {\bf{r}})} \right|}{\|\nabla_{j}(\bf{x}+\bf{r})\|}$, where ${\bf{r}}$ is the current feasible adversarial perturbation. $\|\nabla_{j}(\bf{x}+\bf{r})\|$ denotes the L2 norm of the gradient vector $\frac{\partial{ {h}_j(\hat{\bf x})}}{\partial{\hat{\bf x}}}$ at the point $\hat{\bf x} = \bf{x}+\bf{r}$. 
	\end{lemma}
	
	
	Therefore, instead of running evasion attack for each candidate label, we can simply choose the one with the smallest ratio $\frac{\left| y_{j}{{h_j}({\bf{x}} + {\bf{r}})} \right|}{\|\nabla_{j}(\bf{x}+\bf{r})\|}$. Algorithm \ref{alg:estimator} presents the proposed \emph{Greedy Attack Space Expansion} (GASE) algorithm. It only runs in total \emph{$k-1$} evasion attacks to reach $|S|=k$. 
	
	In the proposed GASE algorithm, the step of \emph{greedy label expansion} is equivalent to conducting the orthogonal matching pursuit guided greedy search \cite{ElenbergAA2016}. It enjoys fast computation, the optimal value of the objective function in Eq.(\ref{eq:bilevelopt}) achieved by GASE has a guaranteed approximation accuracy to the underlying global optimum according to Theorem 1.3 in \cite{Buchbinder2014}.
	
	The step of \emph{greedy label expansion} in Algorithm.\ref{alg:estimator} benefits from label correlation in multi-label instances. A successful attack targeted at one label tends to bias the classification output of another highly correlated label simultaneously. The candidate label  with the weakest classification margin while a large $\|\nabla_{j}(\bf{x}+\bf{r})\|$ is thus likely to be flipped with minor update on the adversarial perturbation. Notably, the proposed GASE algorithm is independent of \emph{the choice of evasion attack methods} in the step \emph{targeted evasion attack}. Once the greedy search for each input instance $\bf{x}$ finishes,  we use the average $|S|$ computed over all $\bf{x}$ as \emph{ the empirical attackability indicator.} A larger average $|S|$ indicates a higher attackability of the targeted multi-label classifier.

	\begin{algorithm}
		
		\DontPrintSemicolon
		\SetAlgoLined
		\caption{Greedy Attack Space Expansion}
		\label{alg:estimator}
		\BlankLine
		\textbf{Input:} Instance example $\bf{x}$, a trained multi-label classifier $h$, perturbation norm budget $\mu_{r}$. \\
		\BlankLine
		\textbf{Output:} The set of attacked labels $S$. \\
		\BlankLine
		Initialize $S$ as an empty set and ${\bf r} = 0$.\\
		\BlankLine
		\While{$|S|<m$ \text{and} ${\left\| \bf{r} \right\|_2} < \mu_{r} $}
		{
			\textbf{Greedy label expansion:} Calculate $d_j$ in Eq.(\ref{select}) for each label $j$  { outside $S$}, where ${h}_j(\bf{x}+\bf{r})$ is the probabilistic classification output of   label  $j$, and $t_j$ is the threshold of label decision;
			\begin{equation}
			\small
			{d_j} = \frac{\left| y_{j}{{h_j}({\bf{x}} + {\bf{r}})} \right|}{\|\nabla_{j}(\bf{x}+\bf{r})\|}.
			\label{select}
			\end{equation}
			\newline Update $S = S \cup j$, where label $j (j \notin S)$ is selected randomly from the labels with the least values of Eq.(\ref{select}). \\
			\BlankLine
			\textbf{Targeted evasion attack:} Solve the targeted evasion attack problem with updated $S$ and get the optimized perturbation $\bf{r}^ *$; Update $\bf{r} = {\bf{r}^ * }$. \\  
		}
	\end{algorithm}
	\vspace{-0.3cm}

	
	
	\section{Experiments}
	In the experimental study, we aim at 1) validating the theoretical attackability analysis  in Theorem 1 and 2; and 2) evaluating the empirical attackability indictor estimated by   GASE     for targeted classifiers. 
	
	
	
	\noindent{\textbf{Datasets.}}  
	We include 4 datasets collected from various real-world  multi-label applications, cyber security practices (\textit{Creepware}),  biology research (\textit{Genbase}) \cite{Tsoumakas2010}, object recognition (\textit{VOC2012}) \cite{pascal-voc-2012} and environment research (\textit{Planet}) \cite{planet2017}. The  4 datasets are summarized in Table.\ref{tab:dataset}. 
	
	
	\noindent{\textbf{Targeted Classifiers.}} 
	We instantiate the study empirically with linear Support Vector Machine (SVM) and Deep Neural Nets (DNN) based multi-label classifiers. 
	Linear SVM is applied on \textit{Creepware} and \textit{Genbase}.     DNN model Inception-V3 is used on \textit{VOC2012} and \textit{Planet}. On each data set, we randomly choose $50\%$, $30\%$ and $20\%$ data instances for training, validation and testing to build the targeted multi-label classifier.  In Table.\ref{tab:dataset}, we show \textit{Micro-F1} and \textit{Macro-F1} scores derived on the unperturbed testing data. {Note that feature engineering and model design of the classifiers for better classification is beyond the scope of this study. These classifiers are trained to achieve comparable classification accuracy w.r.t. the reported state-of-the-art methods on their corresponding datasets, so as to set up the test bed for the attackability analysis.} Due to space limit, more experimental setting and results are provided in the supplementary file. 

	\begin{table}[t!]
		\small                    
		\setlength\tabcolsep{2.0pt} 
		\caption {Summary of the used real-world  datasets. 
			$N$ is the number of instances. $m$ is the total number of labels. $ {l}_{avg}$ is the average number of labels per instance.
			The F1-scores of the targeted classifiers on different datasets are also reported. }	\small 
		\label{tab:dataset}
		\newcommand{\tabincell}[2]{\begin{tabular}{@{}#1@{}}#2\end{tabular}}
		\centering
		\begin{tabular}{c|c|c|c|c|c|c} 
			\hline
			\hline
			Dataset	& $N$ & m & ${l}_{avg}$ & Micro F1 & Macro F1 & Classifier$_{target}$\\
			\hline
			{Creepware}	&966	&16 & 2.07   &0.76  & 0.66   & SVM		\\
			\hline
			{Genbase}	&662 	&27 & 1.25   & 0.99  &0.73  & SVM	  \\
			\hline
			{VOC2012}	&17,125 	&20 & 1.39  & 0.83 &0.74  & Inception-V3  \\
			\hline
			{Planet}	&40,479 	&17 & 2.87   & 0.82  &0.36   & Inception-V3   \\
			\hline
			\hline
		\end{tabular}
	\end{table}
	\vspace{-0.01cm}
	

	\noindent {\textbf{Attack and Adversarial Training.}} \ \ We use  adversarial-robustness-toolbox \cite{art2018toolbox} to implement the step of targeted adversarial attack in Algorithm.\ref{alg:estimator} and adversarial training. Specifically, \textit{projected gradient decent} (PGD) \cite{Madry2018iclr} is employed to conduct the targeted attack in Algorithm.\ref{alg:estimator}. The decision threshold $t_{i}$ in Algorithm.\ref{alg:estimator} is set to 0 without loss of generality. 
	
	
	\noindent \textbf{Performance Benchmark.}\ \ We gradually increase the attack strength by varying the attack budget $\mu_{r}$. Given a fixed value of $\mu_{r}$, we calculate \emph{the average number of flipped labels on test data} as an estimation of the empirical classification risk $R^{emp}_{\mathcal{P}'}$ induced by the attack. This is the empirical attackability indicator, as defined in the end of the section of fast greedy attack space exploration. 

	\subsection{Validation of Empirical Attackability Indicator}\label{sec:empirical_evaluation}
	We assess here  the empirical attackability indicator estimated by the proposed GASE algorithm, by comparing it with   four baselines of label exploration strategies.  
	\begin{itemize}
		\item \textbf{PGS} (Primitive Greedy Search):  {This is the costly primitive greedy search that     requires \emph{$(m+1)k-k(k-1)/2-1$} evasion attacks before including $k$ labels in $S$.} 
		\item \textbf{RS} (Random Search): In each round of RS, one label is selected purely by random from the candidate set without evaluating the marginal gain and added to the current set $S$. 
		\item \textbf{OS} (Oblivious Search): 
		This method first computes the norm of the adversarial perturbation induced by \emph{flipping each candidate label while keeping the other labels unchanged}. The labels causing the least perturbation magnitudes are selected to form the set $S$. 
		\item \textbf{LS} (Loss-guided Search): In each iteration, \textit{LS} updates the adversarial perturbation $\bf r$ along the direction where the multi-label classification loss  increased the most. The set of the attacked labels are reported when ${\left\| \bf{r} \right\|_2}$ surpasses the cost limit. This strategy aims at pushing the originally miss-classified instances even further from the  decision plane, instead of flipping the labels of the originally correctly predicted instances. It misleads the search of the attackable labels by just maximizing the loss, and thus has bad performance as shown in Fig. 1. 
	\end{itemize}
	
	\begin{figure}[t!] 	\label{perrank-all}
		\centering
		\includegraphics[width=0.74\columnwidth]{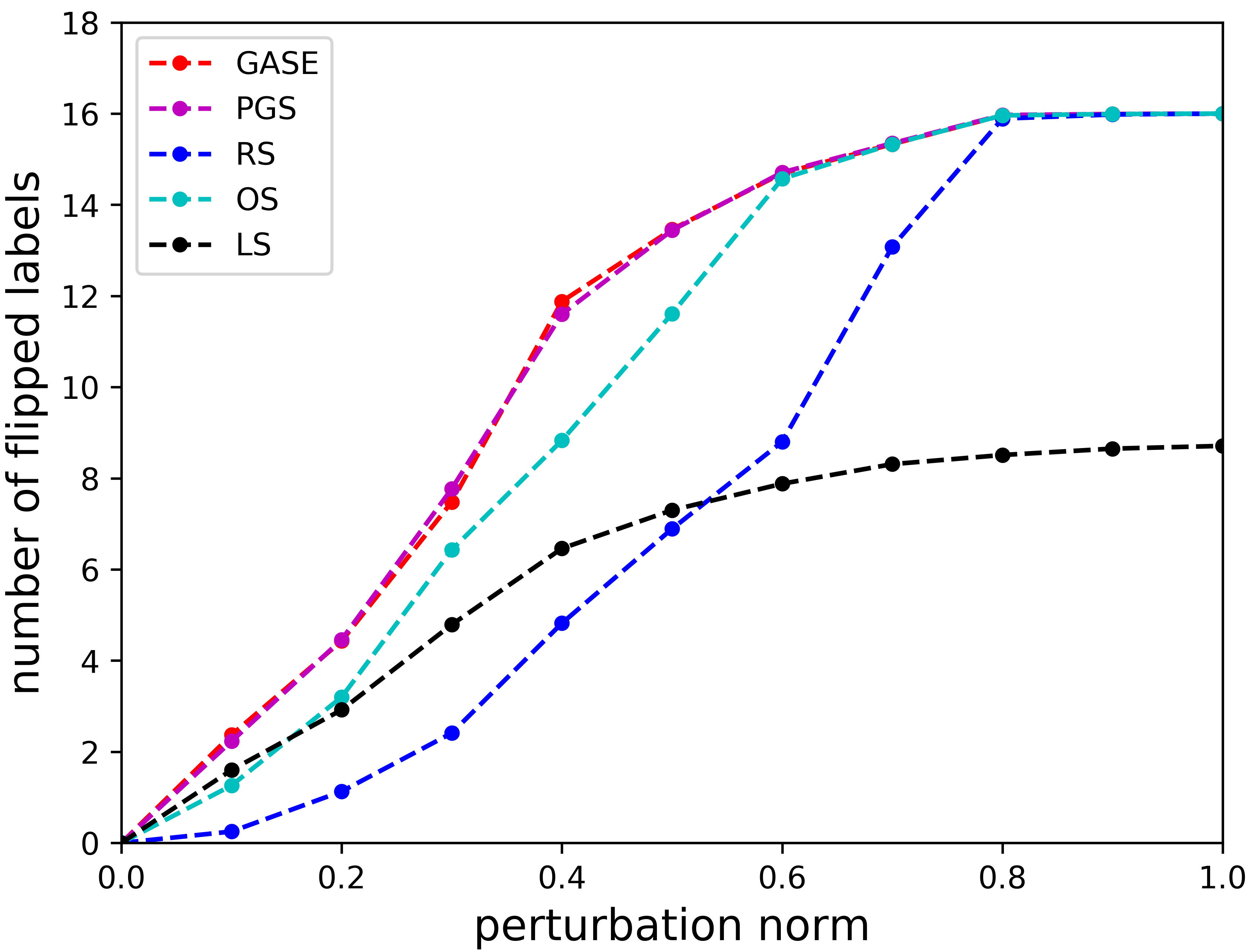}\\
		{\small (a) attackability of SVM on \textit{Creepware}}
		
		\includegraphics[width=0.74\columnwidth]{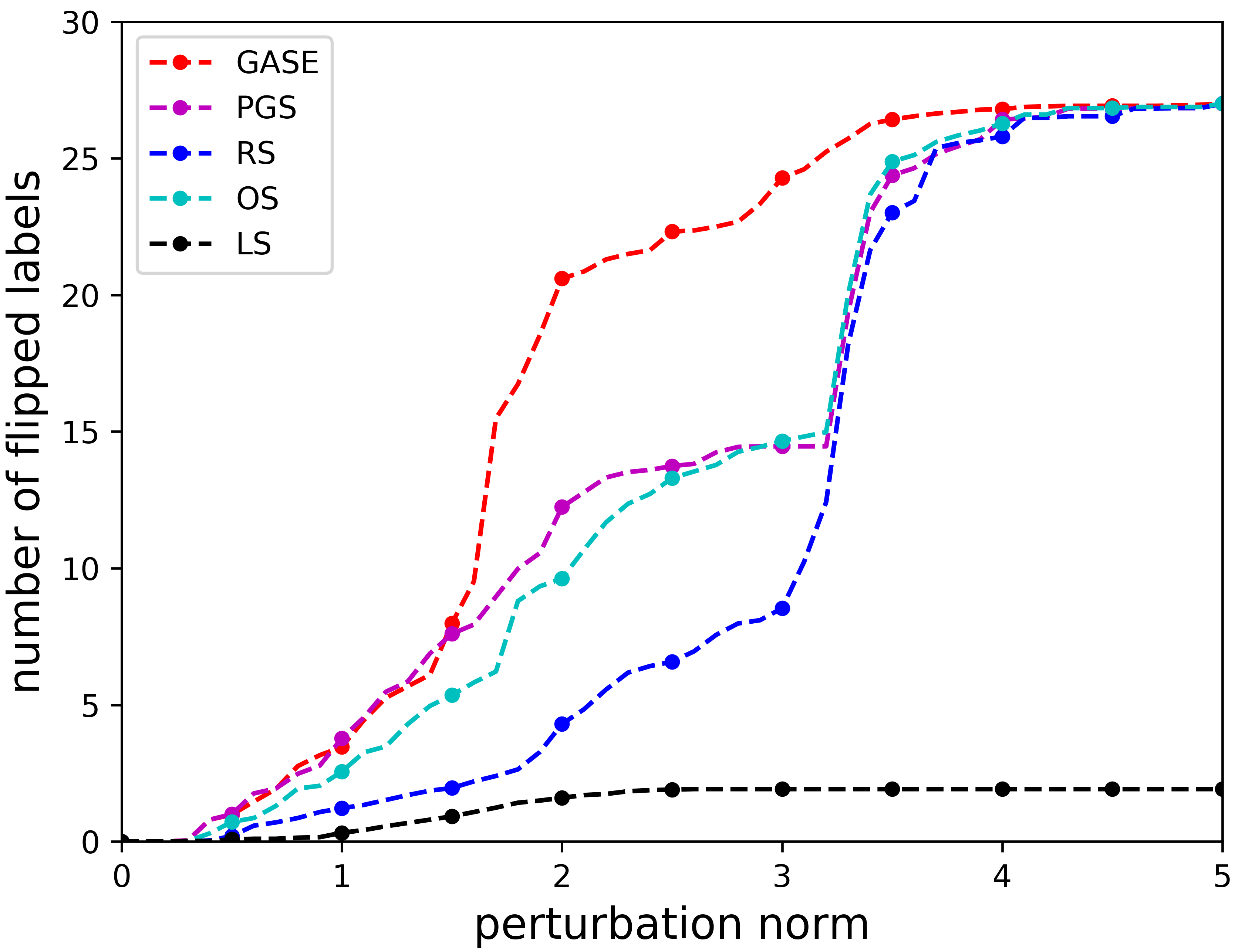}\\
		{\small (b) attackability of SVM on \textit{Genbase}}
		
		\includegraphics[width=0.74\columnwidth]{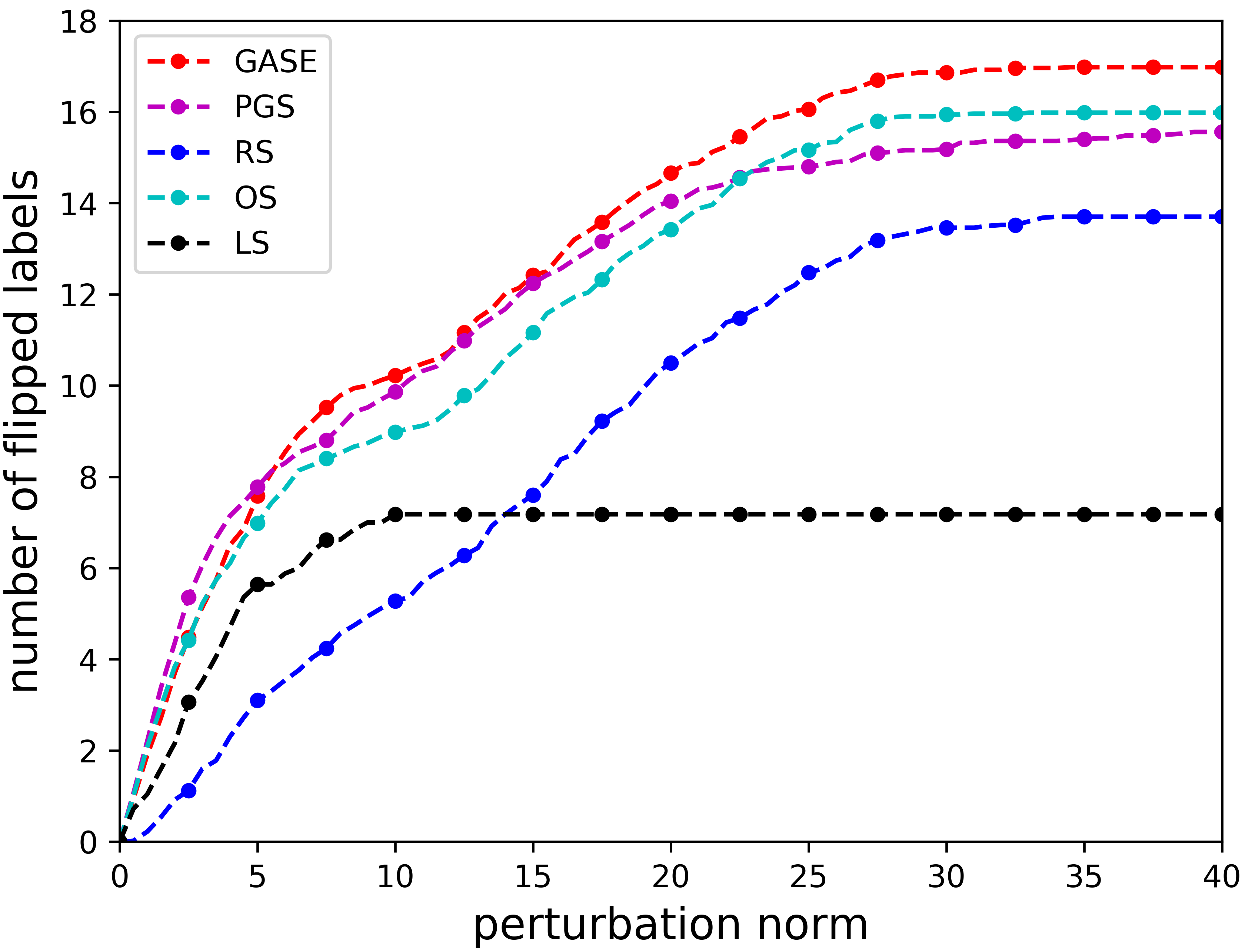}\\
		{\small (c) attackability of Inception-V3 on \textit{VOC2012}}
		\includegraphics[width=0.74\columnwidth]{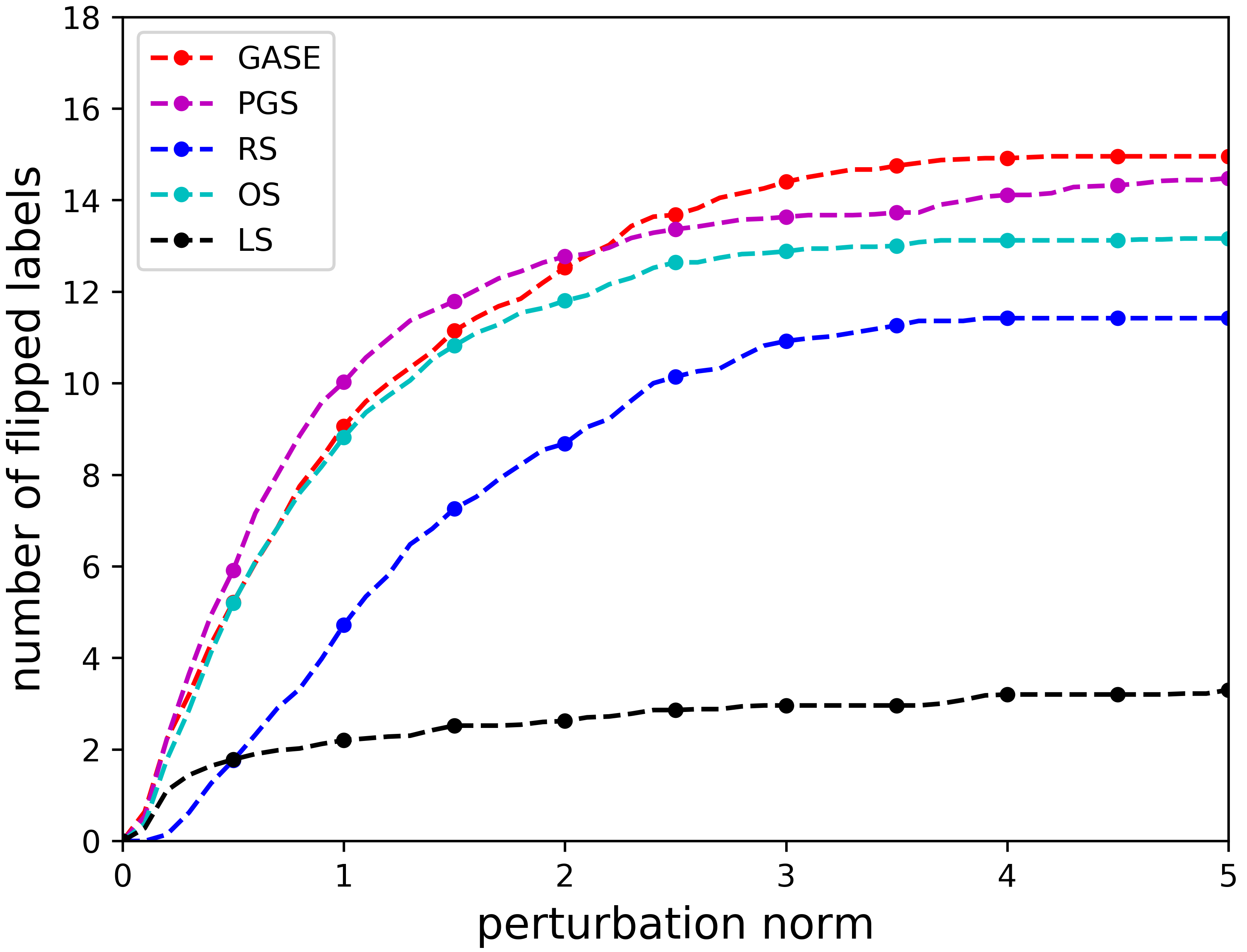}\\
		{\small (d) attackability of Inception-V3 on  \textit{Planet}}
		
		\caption{The empirical attackability indicator estimated by different label exploration strategies. }
	\end{figure} 
	\vspace{-0.1cm}

	\begin{figure}[t!]	\label{permodel-all}
		\centering
		\includegraphics[width=0.77\columnwidth]{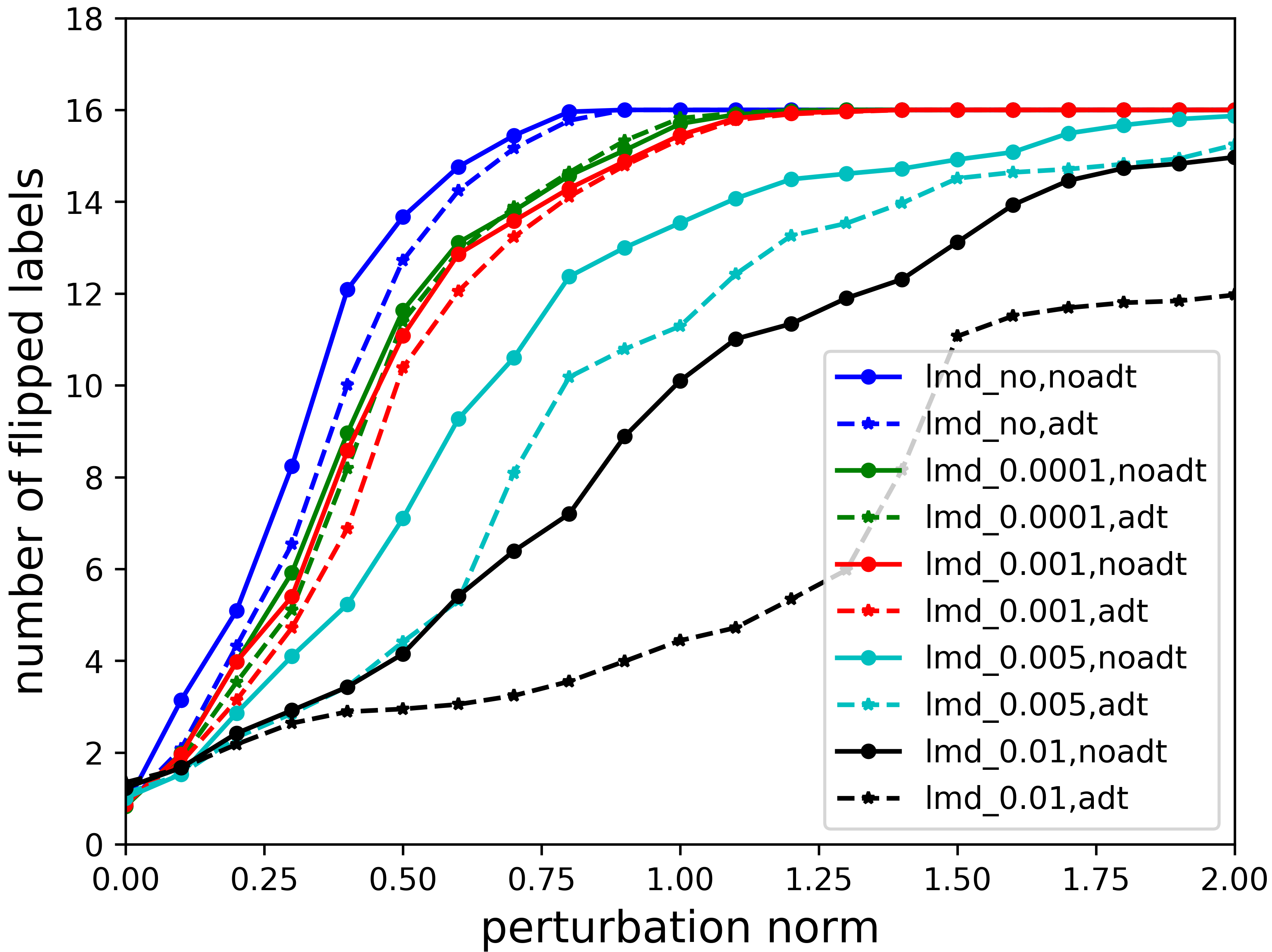}\\
		{\small (a)   attackability  of controlled SVM on  \textit{Creepware}  }
		\includegraphics[width=0.77\columnwidth]{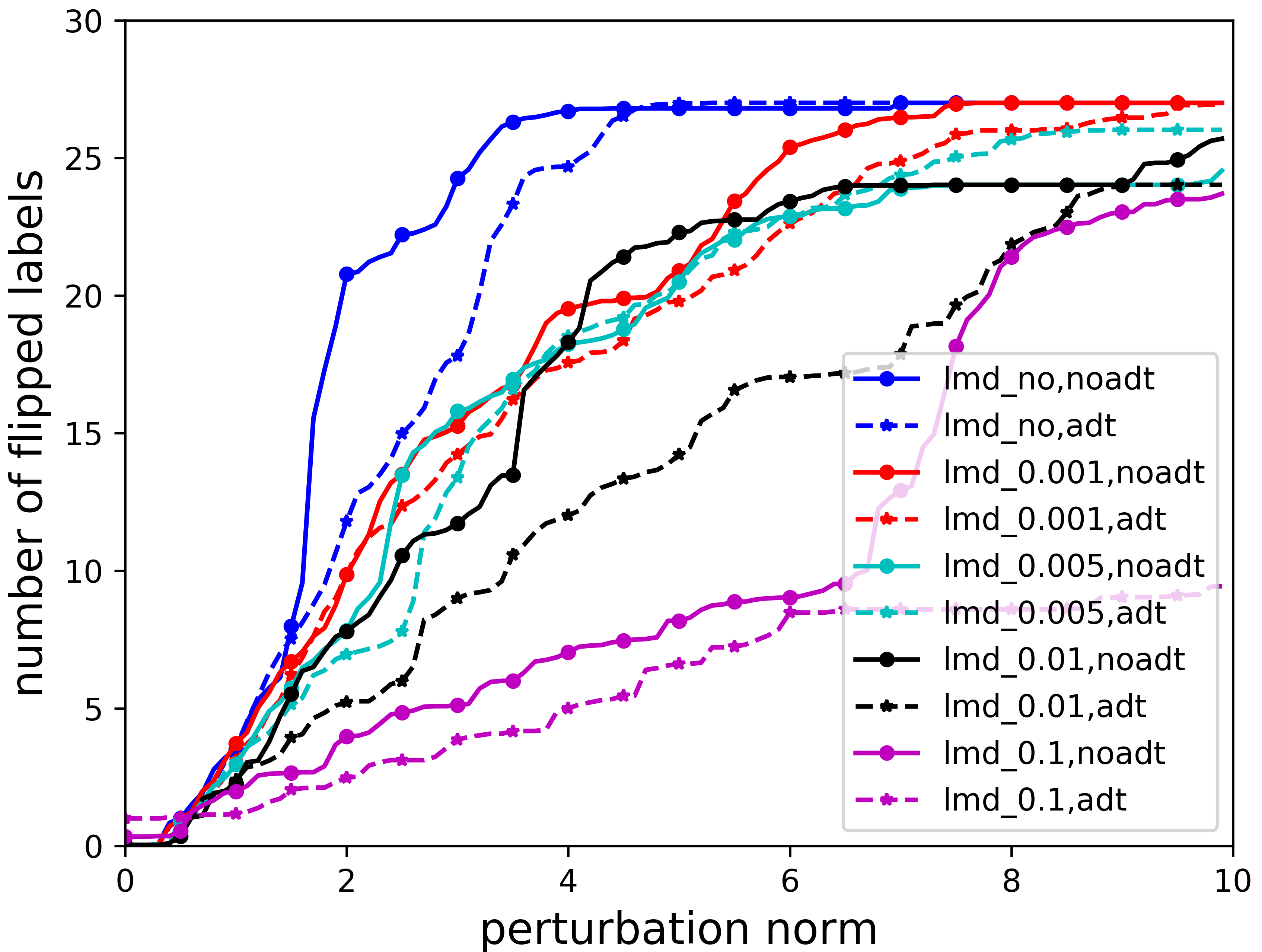}\\
		{\small  (b)   attackability  of controlled SVM on  \textit{Genbase}  }
		\includegraphics[width=0.77\columnwidth]{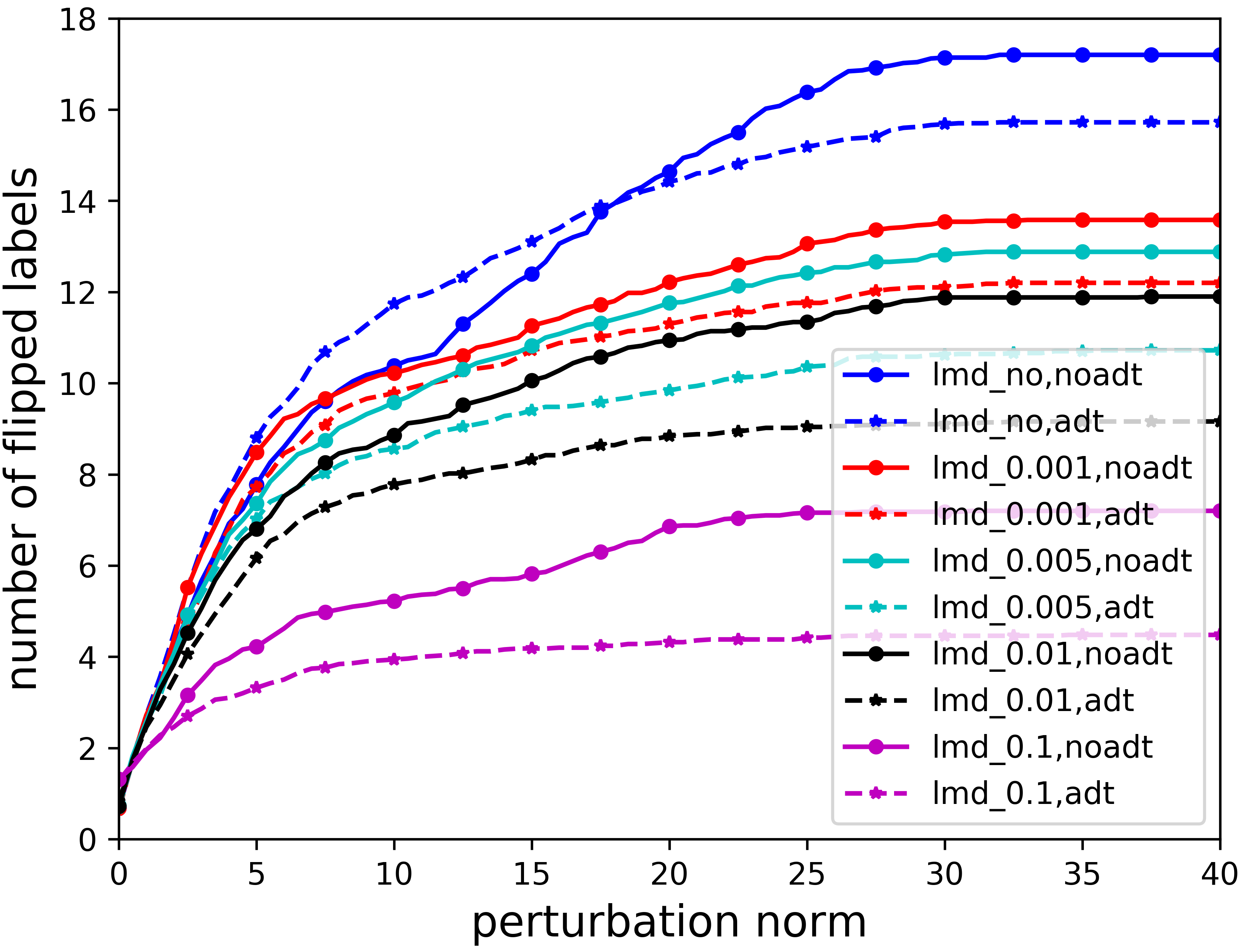}\\
		{\small  (c)   attackability  of controlled Inception-V3 on  \textit{VOC2012}}
		\includegraphics[width=0.77\columnwidth]{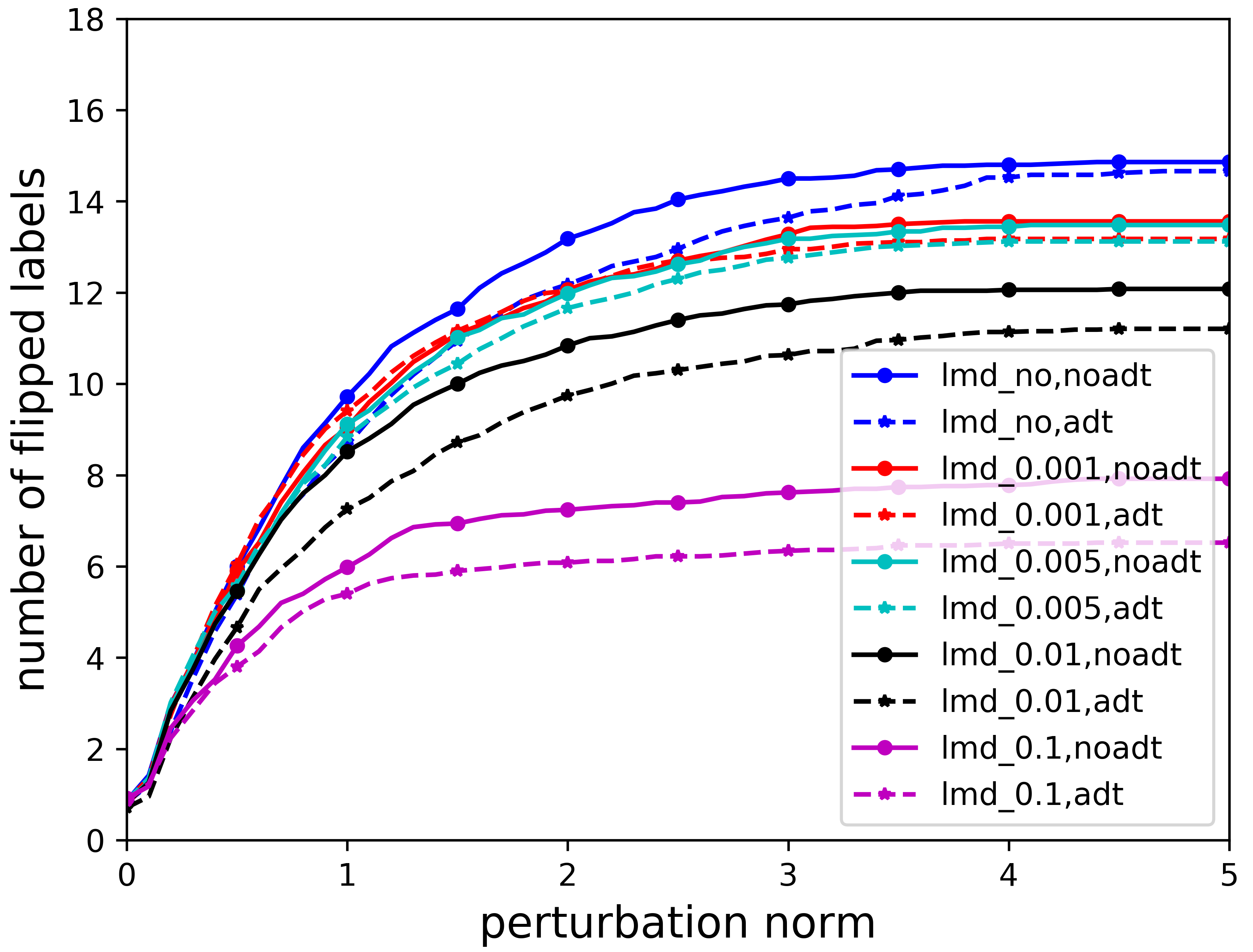} \\
		{\small   (d)   attackability  of controlled Inception-V3 on  \textit{Planet}}
		
		\caption{The  evaluation of classifiers' attackability under different complexity controls. }
	\end{figure}

	Fig. 1 shows the number of flipped labels obtained by the proposed \textit{GASE} algorithm and the baselines on linear and DNN based multi-label classifiers. Since we limit the maximum iterations and perturbation norm bounds of attacks in our experiments, few cases of the involved label exploration methods can flip all of the labels in each dataset. Not surprisingly, the proposed \textit{GASE} and \textit{PGS} method achieve significantly more flipped labels than \textit{RS}, \textit{OS} and \textit{LS} methods, especially when the constraint of attack budget is strict (with small perturbation norms). It confirms the reasonableness of greedy search stated in Lemma.\ref{lemma:submodularity}. Over all the datasets, \textit{GASE} performs similarly or even better compared to \textit{PGS}. It empirically demonstrates the merits of \textit{GASE}: it is much less costly than \textit{PGS}, while obtains   attackability indicators with certified quality. 
	
	\subsection{Attackability Evaluation with Countermeasures for Evasion Attack}\label{sec:adv_risk_mitigation}
	Following in our Theorem 1 and 2, we   study the impact of the countermeasures on multi-label classifiers' attackability: controlling the model complexity by enforcing the low-rank nuclear norm constraint and conducting adversarial training. For the DNN based classifier, we enforce the nuclear norm constraint only on the linear coefficients of the final layer. We include 4 different settings on controlling the model complexity when training classifiers:
	\begin{itemize}
		\item With neither the low-rank nuclear-norm constraint nor   adversarial training over the linear transformation coefficients, noted as ${\rm{lmd\_no}}$ and ${\rm{noadt}}$, respectively.
		\item With both   the nuclear norm constraint and adversarial training, noted as ${\rm{lmd\_{\lambda}}}$ and ${\rm{adt}}$, where $\lambda$ is the regularization parameter of the nuclear norm constraint. 
		\item Without adversarial training while with the nuclear norm constraint, noted as  ${\rm{lmd\_{\lambda}}}$ and ${\rm{noadt}}$, respectively.
		\item With adversarial training while without the nuclear norm constraint, noted as  ${\rm{lmd\_no}}$ and ${\rm{adt}}$, respectively.
	\end{itemize}
	The attackability indicators of all complexity-controlled classifiers are  estimated by the proposed GASE. The results are shown in Fig. 2.
	The figure   also shows robustness evaluation, as a low attackability indicates a high robustness. Consistently found in all datasets,   the low-rank constraint has a significant stronger impact on the classifier's attackability compared to adversarial training, because the variation of $\lambda$ caused larger change among curves in different colors. Classifiers trained with larger $\lambda$ are more robust. Though adversarial training alone doesn't change drastically the  robustness, combined with the low-rank constraint, they can make the classifiers more robust than using solely either one. 
	{The results confirmed our remarks from Theorem 1 and 2.}
	
	
	Our experimental observations show  that adversarial training doesn't change drastically classifiers' performances on unperturbed test data. In contrast, there is indeed an obvious trade-off between improving a classifier's robustness by imposing the nuclear norm constraint and preserving its good utility on unperturbed test data. A strong nuclear norm constraint improves greatly the adversarial robustness. Nevertheless, it also causes accuracy loss to the classifiers.
	More evaluation results about the accuracy of classifiers under complexity control  are  in the supplementary document.
	
	\section{Conclusion}\label{sec:conclusion}
	In this paper, we propose to assess the attackability  of multi-label learning systems under adversarial evasion attack. We theoretically analyze the bound of the expected worst-case risk on adversarial data instances for linear and neural nets based multi-label classifiers. The resultant risk bound is used to evaluate the attackability of the targeted multi-label learning model. We unveil that the attckability depends heavily on 1) the empirical loss on the unperturbed data, 2) the rank of the targeted classifier's linear transformation coefficients and 3) the attack strength. The former two perspectives characterize the attacked multi-label learning task. The latter is decided purely by the adversary. They are the intrinsic cause and external driving force of the adversarial threat. Practically, we propose a greedy-expansion based label space exploration method to provide  the empirical attackability measurement. Enjoying the submdoularity of the label space exploration problem, the empirical attackability evaluation has a certified approximation accuracy to the underlying true value. Our study intrigues the interpretability of adversarial threats of multi-label learning models. The future work will focus on proposing defensive methods for multi-learning systems with provably robustness.  
	
	\clearpage 
	\bibliography{references.bib}
	
	\clearpage
	
	\section{Supplementary}
	We supple the proofs of theorems in our paper, detailed experimental setup information, experiments no Resnet50 based classifiers and case study by CW attack.
	\section{Proof of Theorem 1}
	We assume that the linear multi-label classifier $h(x) = wx$, where $y\in{R^{m}}$ and $x\in{R^{d}}$ are the label and feature vector of one given instance $z=(x,y)$ respectively.  $w\in{R^{m*d}}$ denotes the linear transformation coefficient matrix of the classifier $h$. We require $\|w\|_{\sigma}\leq{\Lambda}$ ($\|\|_{\sigma}$ is the spectral norm of the coefficient matrix $w$). 
	
	The least-squared error (LSE) risk function that are popularly used in multi-label classification can be formulated as $\ell(x,y) = \|y - wx\|_{2}$. $\|\|_{2}$ denotes Euclidean norm. We define the distance metric in the joint space $\mathcal{Z}=\mathcal{X}\times{\mathcal{Y}}$ as:
	\begin{equation}\label{distance_mllinear_supp}
	d(Z,Z') = \|x-x'\|_{2} + \|y-y'\|_{2}.
	\end{equation}
	For any $x\in{\mathcal{X}}$ in the feature space, we require that $\|x\|_{2}\leq{\mu_{x}}$. Therefore we have $d(z,z')\leq{2\mu_{x}+l}$ for any $z$ and $z'$ in the joint space $\mathcal{Z}$. The upper bound of $f$ is given by $0\leq{f}\leq{M}$,where $M=m+\Lambda{\mu_{x}}$.

	
	Further we have: for any $z$ and $z'$ in the joint space $\mathcal{Z}$, we have
	\begin{equation}
	\begin{split}
	|\ell(z') - \ell(z)| &\leq |\|wx'-y'\|_2 - \|wx-y\|_2| \\
	&\leq \|w(x'-x)\|_2 + \|y'-y\|_2\\
	&\leq \|w\|_{\delta}\|x'-x\|_2 + \|y'-y\|_2\\
	= C_{h}d(z',z),
	\end{split}
	\end{equation}
	where $C_{h} = max\{\|w\|_{\sigma},1\}$.
	
	\begin{lemma}\label{lemma:operator}
		Let $S$:$X\longrightarrow{Y}$ be the operators in real Banach space and $\epsilon >0$. Then the covering number $\mathcal{N}$ of $T$ can be bounded as: 
		\begin{equation}
		\mathcal{N}(T,\|\|,\epsilon) \leq (1 + \frac{2\|T\|}{\epsilon})^{R},
		\end{equation}
		where $R$ is the rank of the operator $T$ and $\|\|$ is the norm of the operator. 
	\end{lemma}
	The proof of the Lemma follows the same lines as the proofs for the similar properties of entropy numbers (Section.1d)\cite{konig1986}.

	Our analysis is conducted based on combining together Lemma.2 in Section.3 and Lemma.6 in Section.4 in \cite{TuNIPS19}, which gives:
	\begin{lemma}
		Let $R_{\mathcal{P}'}(h)$ and $R^{emp}_{\mathcal{P}'}(h)$ denote the expected and empirical worst-case risk under the evasion attack, as defined in Definition.1. We have the upper bound of $R_{\mathcal{P}'}(h)$ holds with probability at least $1-\sigma$:
		\begin{equation}\label{eq:dudley_upperbound}
		\begin{split}
		&R_{\mathcal{P}'}(h) \leq R^{emp}_{\mathcal{P}'}(h) + \frac{24\kappa}{\sqrt{n}} \\
		&+ M\sqrt{\frac{\log(1/\sigma)}{2n}} + \frac{12\sqrt{\pi}}{\sqrt{n}}{C_{h}}diam(Z),\\
		&\kappa = \int_{0}^{\infty}{\sqrt{\log{\mathcal{N}(\mathcal{F},\|\|_{\infty},u/2)}}}du,\\
		&R^{emp}_{\mathcal{P}'}(h) \leq \frac{1}{n}\sum_{i=1}^{n}\ell({\bf x}_i,{\bf y}_i) + C_{h}\mu_{r},\\
		\end{split}
		\end{equation}
		where $\mathcal{N}$ denotes the covering number of the functional $\mathcal{F}$, where the loss function $f\in{\mathcal{F}}$. $diam(Z)$ denotes the diameter of the $L2$-ball of the joint space $\mathcal{Z}$. $\mu_{r}$ is the limit of the attack budget. 
	\end{lemma}
	The proof of this lemma can be derived by combining the conclusion of Lemma.2 in Section.3 and Lemma.6 in Section 4 in \cite{TuNIPS19}. Furthermore, we bound $\Lambda_{\epsilon_{\mathcal{B}}}$ in Lemma.6 in Section.4 with $C_{h}$ in our study to indicate the impact of the spectrum of $w$ over the expected worst-case risk bound. 
	
	To evaluate dudley entropy integral in Eq.\ref{eq:dudley_upperbound} to compute the covering number $\mathcal{N}(\mathcal{F},\|\|_{\infty},u/2)$, we give:
	\begin{equation}\label{eq:inf_upperbound}
	\begin{split}
	\|\ell-\ell^{'}\|_{\infty}&=\sup_{x\in{\mathcal{X}},y\in{\mathcal{Y}}}|\|wx-y\|_{2} - \|w'x-y\|_{2}|\\
	&\leq \|(w-w')x|\|_{2}\\
	&\leq \mu_{x}\|w-w'\|_{\sigma}\\
	&\leq 2\mu_{x}\|w\|_{\sigma}.\\
	\end{split}
	\end{equation}
	Therefore the covering number $\mathcal{N}$ is bounded base on Lemma.\ref{lemma:operator}:
	\begin{equation}
	\mathcal{N}(\mathcal{F},\|\|_{\infty},u/2)\leq (1+\frac{4\mu_{x}\Lambda}{u})^{R}.
	\end{equation}
	$\Lambda$ bounds the $\|w\|_{\sigma}$. $R$ is the rank of the operator for classification risk calculation $\ell:\mathcal{X}\times\mathcal{Y}\longrightarrow{R}$. We can formulate $f$ with matrix transformation:
	\begin{equation}
	\begin{split}
	h(x,y) &= wx-y = [w,-\mathbf{1}][x^{T},y^{T}]^{T}\\
	\ell(x,y) &= h(x,y)^{T}h(x,y).\\
	\end{split}
	\end{equation}
	Therefore $R$ is no more than the rank of $w$. We use $R$ to denote the rank of $w$ hereafter. 
	
	For $u{\geq}2{\mu_{x}}\Lambda$, $\mathcal{N}(\mathcal{F},\|\|_{\infty},u/2) = 1$. We can hence absorb some positive multiplicative constants into $\Lambda$ and formulate the upper bound of the covering number as follows:
	
	\begin{equation}\label{eq:covnum_upperbound}
	\begin{split}
	&\int_{0}^{\infty}\sqrt{\log\mathcal{N}(\mathcal{F},\|\|_{\infty},u/2)}du \leq \int_{0}^{2\mu_{x}\Lambda}\sqrt{\log{R(1+\frac{4\mu_{x}\Lambda}{u})}}du\\
	&\leq 2\sqrt{R(2+2\mu_{x}\Lambda)}\sqrt{2\mu_{x}\Lambda}.\\
	\end{split}
	\end{equation}
	
	Substituting this into Eq.\ref{eq:dudley_upperbound}, we get the desired result in Theorem.1: 
	
	\begin{equation}
	\begin{split}
	&R_{\mathcal{P}'}(h) \leq R^{emp}_{\mathcal{P}'}(h) +
	96\sqrt{\frac{\mu_{x}{\Lambda}R(1+\mu_{x}\Lambda)}{n}} \\
	&+ \frac{12{C_{h}}\sqrt{\pi}(m+2\mu_{x})}{\sqrt{n}} + (m+\Lambda{\mu_{x}})\sqrt{\frac{\log(1/\sigma)}{2n}}. \\
	\end{split}
	\end{equation}

	In the adversary-free learning scenario, we can derive the generalization error bound of $h$ as follows: 
	
	\begin{corollary}\label{corollary_linear}
		Given the setting of the multi-label data and the linear multi-label classifier $h$, the upper bound of the expected generalization error of $h$ without the adversary holds with at least probability of $1-\sigma$ as follows: 
		\begin{equation}\label{eq:riskbound_linear_advfree_supp}
		\begin{split}
		&R_{\mathcal{P}}(h)\leq {\frac{1}{n}\sum_{i=1}^{n}\ell(\bf{x}_i,\bf{y}_i)} + 96\sqrt{\frac{\mu_{x}{\Lambda}R(1+\mu_{x}\Lambda)}{n}} \\
		&+ \frac{12{C_{h}}\sqrt{\pi}(m+2\mu_{x})}{\sqrt{n}} +(m+\Lambda{\mu_{x}})\sqrt{\frac{\log(1/\sigma)}{2n}},\\
		\end{split}
		\end{equation}
		where $R_{\mathcal{P}}(h)$ is the expected and empirical risk of $h$ on the distribution $\mathcal{P}$ of multi-label data instances. $R^{emp}_{\mathcal{P}}(h) = {\frac{1}{n}\sum_{i=1}^{n}\ell(\bf{x}_i,\bf{y}_i)}$ denotes the empirical classification risk of $h$ over the multi-label instances sampled from $\mathcal{P}$. 
	\end{corollary}
	
	\begin{remark}\label{remark2}
		Compare the derived adversary-free risk in Eq.\ref{eq:riskbound_linear_advfree_supp} and the expected risk under adversarial perturbation in Eq.\ref{eq:upperboundlinear}, the extra effect that the adversary introduces is related to the magnitude of the adversarial perturbation $\|\mu_{r}\|$. The attackability depends heavily on the intrinsic regularity of the classifier's architecture and training data distribution. Notably, even in adversary-free scenario, a low-rank structured linear multi-label classifier tends to have lower generalization error.   
		Without the presence of the adversary, all the terms involving the attack budget $\mu_{r}$ vanish.
		According to Eq.\ref{eq:riskbound_linear_advfree_supp}, the classifier $h$ with a lower-rank structure tends to have lower generalization error over the distribution $\mathcal{P}$. This is consistent with the observation reported in previous multi-label research efforts.  
	\end{remark}
	
	\section{Proof of Theorem 2}
	
	Inherited the setting of the attack scenario from Theorem.\ref{theorem:linear}, we consider a neural network based multi-label classifier $h_{nn}$ with $L$ layers, where:
	\begin{itemize}
		\item The dimension of each layer is $d_{1},d_{2},...,d_{L}$, and $d_{0} = d$ for taking input $\bf x$ 
		and $d_{L} = m$ for outputting labels $\bf y$ .  
		\item At each layer $i$, $A_{i}\in{R^{d_{i-1}{\times}d_{i}}}$ denotes the linear coefficient matrix (connecting weights). The spectral norm of $A_{i}$ is bounded as $\|A_{i}\|_{\delta}\leq{\Lambda_{i}}$. $R_{i}$ denotes the rank of $A_{i}$. 
		\item The activation functions used in the same layer are Lipschitz continuous and bounded. We assume that the activation functions used in the same layer share the same Lipschitz constant $\rho_{i}$. We use $g_{i}$ to denote the activation functions used at the layer $i$. The output of each layer $i$ can be defined recursively as $\mathcal{H}_{i} = g_{i}(\mathcal{H}_{i-1}A_{i})$.
	\end{itemize}
	
	Again we use the least-squared error (LSE) risk function as $\ell(x,y) = \|y - wx\|_{2}$. $\|\|_{2}$ denotes Euclidean norm. We define the distance metric in the joint space $\mathcal{Z}=\mathcal{X}\times{\mathcal{Y}}$ as:
	\begin{equation}
	d(Z,Z') = \|x-x'\|_{2} + \|y-y'\|_{2}.
	\end{equation}
	For any $x\in{\mathcal{X}}$ in the feature space, we require that $\|x\|_{2}\leq{\mu_{x}}$. Therefore we have $d(z,z')\leq{2\mu_{x}+m}$ for any $z$ and $z'$ in the joint space $\mathcal{Z}$. 
	
	
	We observe that the following inequality holds:
	\begin{equation}
	\begin{split}
	\|\ell(x,y) - \ell(x',y')\|_2 &\leq \|y-y'\|_2 + \|\mathcal{H}(x) - \mathcal{H}(x')\|_2,\\
	\|\ell(x,y) - \ell(x',y')\|_2 &\leq \|y-y'\|_2 + \prod_{j=1}^{L}\rho_{j}\|A_j\|_{\delta}\|x-x'\|_{2},\\
	\|\ell(x,y) - \ell(x',y')\|_2 &\leq C_{nn} d_{Z}(z,z'),\\
	\end{split}
	\end{equation}
	where $C_{nn} = max\{1,\prod_{j=1}^{L}\rho_{j}\|A_j\|_{\delta}\}$. 
	
	In the followings, we prepare to compute the covering number of $h_{nn} \in \mathcal{F}$. First for any two feed-forwrd neural network $\mathcal{H}_{\mathcal{A}}$ and $\mathcal{H}_{\mathcal{A}'}$, where $A=(A_1,A_2,A_3,...,A_L)$ and $A'=(A'_1,A'_2,A'_3,...,A'_L)$, we have the following bounds:
	\begin{equation}\label{eq:inftynorm}
	\begin{split}
	&\|\ell-\ell'\|_{\infty}=\sup_{x\in{\mathcal{X}},y\in{\mathcal{Y}}}|\|\delta_{L}(\mathcal{H}_{L-1}(x)A_{L})-y\|_{2} -\\ &\|\delta_{L}(\mathcal{H}'_{L-1}(x)A'_{L})-y\|_{2}|\\
	&\leq \sum_{i=1}^{L}C_{i}\|\mathcal{H}_{L-i}(x)(A_{L+1-i} - A'_{L+1-i})\|_{2},\\
	\end{split}
	\end{equation}
	where $C_{1} = \rho_{L}$ and $C_{i} = \prod_{j=L-i+1}^{L}\rho_{j}\prod_{j=L+2-i}^{L}\|A_{j}\|_{\delta}$ with ${L}\geq{i}\geq{2}$, 
	
	Based on the definition of the covering number,  Eq.\ref{eq:inftynorm} and Lemma.\ref{lemma:operator}, we derive the covering number of $h_{nn}$ as:
	\begin{equation}
	\log(\mathcal{N}(\mathcal{F},\|\|_{\infty},u/2))\leq \sum_{i=1}^{L} R_{i}\log(1+4L\frac{B_{i}C_{i}\Lambda_{i}}{u}),     
	\end{equation}
	where $R_{i}$ denotes the rank of $A_{i}$. $\|\mathcal{H}_i(x)\|_{2}\leq{B_{i}}$. Since the activation function of each layer is bounded, we can further assume that the activation function is Sigmoid or Tanh function. In this case, $\|\mathcal{H}_i(x)\|_{2}\leq{d_{i}}$. 
	
	Therefore the upper bound of the dudley entropy integral gives: 
	\begin{equation}
	\begin{split}
	&\int_{0}^{\infty}\sqrt{\log(\mathcal{N}(\mathcal{F},\|\|_{\infty},u/2))}du \\
	&\leq \sum_{i=1}^{L}\int_{0}^{2d_{i}\Lambda_{i}}\sqrt{ R_{i}\log(1+4L\frac{d_{i}C_{i}\Lambda_{i}}{u})}{2LC_{i}}du_{i}\\ 
	&\leq 4\sqrt{mL\Lambda_{L}}\sum_{i=1}^{L}R_i\sqrt{d_i{\Lambda_{i}}C_i},\\
	\end{split}
	\end{equation}
	where we absorb some positive multiplicative constants into $\Lambda_{i}$. We finally find that the expected adversarial risk bound of the feed-forward neural network model holds with the probability no less than $1-\delta$ as:
	
	\begin{equation}\label{eq:upperboundnn_supp}
	\small
	\begin{split}
	&R_{\mathcal{P}'}(h_{nn})\leq R^{emp}_{\mathcal{P}'}(h_{nn}) + 2m\sqrt{\frac{\log(1/\sigma)}{2n}} + \\
	&\frac{96\sqrt{mL\Lambda_L}\sum_{i=1}^{L}{R_i}\sqrt{d_{i}{\Lambda_i}C_i}}{\sqrt{n}} + \frac{12{C_{nn}}(2\mu_{x}+m)\sqrt{\pi}}{\sqrt{n}},  
	\end{split}
	\end{equation}
	and the empirical loss $R^{emp}_{\mathcal{P}'}$ has the upper bound: 
	\begin{equation}\label{eq:upperboundnn1_supp}
	\small
	\centering
	R^{emp}_{\mathcal{P}'}(h_{nn}) \leq \frac{1}{n}\sum_{i=1}^{n}\ell({\bf x}_i,{\bf y}_i) + C_{nn}\mu_{r}.
	\end{equation}

	We include in a further step the adversary-free generalization bound of $h_{nn}$ based on Eq.\ref{eq:upperboundnn_supp}: 
	
	\begin{equation}\label{eq:upperboundnn_adv_free_supp}
	\small
	\begin{split}
	&R_{\mathcal{P}}(h_{nn})\leq \frac{1}{n}\sum_{i=1}^{n}\ell({\bf x}_i,{\bf y}_i) + 2m\sqrt{\frac{\log(1/\sigma)}{2n}} + \\
	&\frac{96\sqrt{mL\Lambda_L}\sum_{i=1}^{L}{R_i}\sqrt{d_{i}{\Lambda_i}C_i}}{\sqrt{n}} + \frac{12{C_{nn}}(2\mu_{x}+m)\sqrt{\pi}}{\sqrt{n}}.   
	\end{split}
	\end{equation}
	
	\begin{remark}
		The generalization risk bound given in Eq.\ref{eq:upperboundnn_adv_free_supp} illustrates explicitly the association between the low rank linear transformation coefficients in the neural network based classifier and its generalization capability for multi-label classification. As unveiled, at least one layer of the neural network should be of a low-rank structure, in order to improve its expected classification accuracy over unknown testing samples. Furthermore, lower spectral norm (smaller leading eigenvalues) of the linear coefficients can also improve the generalization capability and robustness under the evasion attack. The analysis also unveils the relation between generalization capability and adversarial robustness of a multi-label classifier. 
		
		Notably, \cite{TuNIPS19} gave an adversarial risk bound of a feed-forward neural network model used for binary classification. However, in multi-label classification scenarios, we are especially curious \emph{whether a multi-label classifier can gain adversarial robustness from the low rank constraint}. The attackability analysis in our work thus provides an explicit answer to the question.

	\end{remark}
	
	
	\section{Proof of Lemma 1 and Lemma 2}
	We first verify the supermodularity of $g(S)$ in Eq.\ref{eq:bilevelopt}. $g(S)$ is a non-decreasing set function with increasingly larger $S$. We first derive an analytical solution to $g(S)$ with lagrangian multipliers $\{\lambda_{i}\}$:
	\begin{equation}\label{eq:lpcp}
	\begin{split}
	&J(\lambda_{i},r) = \|r\|^2 \\
	&+ \sum_{i=1}^{m}\lambda_{i}(2{b}_{i}y_i{h_{i}(x+r)} - y_{i}h_{i}(x+r)+t_i),\\
	s.t. &\,\,\,\lambda_{i}\geq{0},\\
	&b_{i} = 1 \,\,\,for\,\,\,i\in{T},\\
	&b_{i} = 0 \,\,\,for\,\,\,i\notin{T},\\
	\end{split}
	\end{equation}
	where $h_{i}$ denotes the output of the classifier $h$ corresponding to the $i$-th label. Since the adversarial noise $r$ is usually of small magnitude, we further approximate $h_{i}(x+r)$ with its Taylor expansion: $h_{i}(x+r) \approx h_{i}(x) + r^{T}h'_{i}(x)$. Eq.\ref{eq:lpcp} can be further simplified as a quadratic programming problem with affine constraints:
	
	\begin{equation}\label{eq:lpcp1}
	\begin{split}
	&J(\lambda_{i},r) = \|r\|^2 \\
	&+ \sum_{i=1}^{m}\lambda_{i}(2{b}_{i}y_{i}(h_i+r^{T}\phi_{i}) +t_i - y_{i}h_{i} - y_{i}r^{T}\phi_{i}),\\
	s.t. &\,\,\,\lambda_{i}\geq{0},\\
	&b_{i} = 1 \,\,\,for\,\,\,i\in{T},\\
	&b_{i} = 0 \,\,\,for\,\,\,i\notin{T},\\
	\end{split}
	\end{equation}
	where $\phi_{i}(x) = h'_{i}(x)$ denotes the gradient of $h_{i}(x) $ with respect to the input feature vector $x$. 
	
	By taking the first-order condition $\frac{\partial{J}}{\partial{r}} = 0$, we can derive the optimal $\|r\|^2$ as :
	\begin{equation}\label{eq:rform}
	\|r\|^2 = \frac{1}{4} \|\sum_{i=1}^{m}(\lambda_{i}y_{i}\phi_{i} - 2\lambda_{i}\phi_{i}b_{i}y_{i})\|^{2}_{2},\\
	\end{equation}
	
	and according to KKT conditions, we can get for non-zero $\lambda_{i}$: 
	
	\begin{equation}
	\begin{split}
	&h_{i} + \frac{1}{2}\sum_{k\in{\{k|\lambda_{k}>0\}}}(\lambda_{k}y_{k}\phi^{T}_{k}\phi_{i} - 2\lambda_{k}\phi^{T}_{k}\phi_{i}\hat{y}_{k}) = 0 \\
	&(if\,\,\lambda_{i}>0).\\ 
	\end{split}
	\end{equation}
	
	\begin{observation}
		To obtain the values of $\lambda_{k}>0$ ($k=0,1,2,...,K$), we turn to solve the equation system such as:
		\begin{equation}\label{eq:eqset}
		\begin{split}
		&-\frac{1}{2}\Pi[\Phi^{T}_{0},\Phi^{T}_{1},...,\Phi^{T}_{K}] = H,\\
		&\Pi = [\lambda_{0}\phi^{T}_{0}-2b_{0}y_{0}\phi^{T}_{0},\lambda_{1}\phi^{T}_{1}-2b_{1}y_{1}\phi^{T}_{1},\\
		&...,\lambda_{K}\phi^{T}_{K}-2b_{K}y_{K}\phi^{T}_{K}],\\
		&\Phi_{k} = [\phi_{k},\phi_{k},...,\phi_{k}],\\
		&H = [h_{0},h_{1},h_{2},...,h_{K}].\\
		\end{split}
		\end{equation}
		Given a set of $\{b_{0},b_{1},...,b_{K}\}$, the value of $\lambda_{k}$ ($k=0,1,2,...,K$) is determined uniquely by $h_{i}$ and $\phi_{i}$.  
	\end{observation}

	\begin{observation}
		$\|r\|^2$ is a set function defined over the set $T$, as $\hat{y}_{i}$ is determined by the binary variable $b_{i}$. 
	\end{observation}
	\begin{observation}
		$\|r\|^2$ is a convex quadratic function with respect to the variable $\{\hat{y}_{i}\},(i=1,2,3,...,m)$, given a set of $\{b_{0},b_{1},b_{2},...b_{K}\}$ fixed in Eq.\ref{eq:rform}. 
	\end{observation}
	
	Furthermore, by simply flipping the sign of $g(S)$, we can find that $-g(S) = \underset{S}{\min}-\|r\|^2$ is non-increasing and submodular function, since $-\|r\|^2$ is concave, according to Theorem.1 in \cite{ElenbergAA2016}. Correspondingly, $g(S)$ is a non-decreasing supermodular set function. In Eq.\ref{eq:bilevelopt}, $|S|$ is a monotonically increasing modular function. As a result, the objective $\psi(S) = |S| - g(S)$ of the maximization problem defined in Eq.\ref{eq:bilevelopt} is a non-monotone submodular function. According to Theorem 1.5 in \cite{Buchbinder2014}, randomized greedy forward expansion of the set $S$ can provide a guarantee to the approximation accuracy: 
	\begin{equation}
	\psi(\hat{S}) \geq \frac{1}{4}\psi(S^{*}).
	\end{equation}
	where $\psi({\hat{S}})$ and $\psi({S^{*}})$ denote respectively the objective function value obtained by randomized greedy forward search proposed in \cite{Buchbinder2014} and the underlying global optimum following the cardinality lower bound constraint.

	\subsection{Proof of Lemma 2}
	
	In each iteration of the greedy forward search, the current set of flipped labels is noted as $T$ and the current perturbed input is noted as $\tilde{x}$. $h(\tilde{x})$ and $\phi(\tilde{x})$ denotes the current classifier output and gradient vector with respect to $\tilde{x}$. We further assume that the $i^{*}$-th label is selected to be flipped and added to the set $T$ in the current iteration of the greedy search. Given the $h$, we inject a small adversarial perturbation $r$ to form the perturbed input $\tilde{x}+r$ centering at $\tilde{x}$, in order to flip the label $i^{*}$. By taking $\frac{\partial{J}}{\partial{r}} = 0$ and $\frac{\partial{J}}{\partial{\lambda_{i}}} = 0$ and the complementary slackness conditions, we have: 
	\begin{equation}
	\begin{split}
	&\|r_{-i^{*}}\|_{2} \leq \frac{1}{2}\|\sum_{j\neq{i}}^{m}(\lambda_{j}y_{j}\phi_{j}-2\lambda_{j}\phi_{j}\hat{y}_{j})\|_{2}\\
	&+ \frac{1}{2}\|\lambda_{i^{*}}y_{i^{*}}\phi_{i^{*}}\|_{2},\\
	&y_{i}h_{i} + t_{i} \\
	&= -\frac{y_{i}}{2}(\sum_{j\neq{i}}(\lambda_{j}y_{j}\phi^{T}_{j}\phi_{i}-2\hat{y}_{j}\lambda_{j}\phi^{T}_{j}\phi_{i})) + \frac{\lambda_{i}\|\phi_{i}\|^2}{2},\,\,\,\,i\in{T},\\
	&-y_{i}h_{i} + t_{i} \\
	&= \frac{y_{i}}{2}(\sum_{j\neq{i}}(\lambda_{j}y_{j}\phi^{T}_{j}\phi_{i}-2\hat{y}_{j}\lambda_{j}\phi^{T}_{j}\phi_{i})) - \frac{\lambda_{i}\|\phi_{i}\|^2}{2},\,\,\,\,i\notin{T},\\
	&\lambda_{i}(2b_{i}y_{i}h_{i} - y_{i}h_{i} + t_{i} + (2b_{i}y_{i}-y_{i})r^{T}\phi_{i}) = 0,\\
	\end{split}
	\end{equation}
	where $r_{-i^{*}}$ denotes the adversarial perturbation that can flip both the labels in the current attacked label set $T$ and the latest added label $i^{*}$. 
	\begin{observation}
		To minimize $\|r_{-i^{*}}\|_{2}$, we can set $\lambda_{j} = 0$, ($j\neq{i^{*}}$) and $\lambda_{i^{*}} > 0$. In this case, we can derive a feasible solution:
		\begin{equation}\label{eq:optimalr}
		r_{-i^{*}} = -\frac{1}{2}\lambda_{i^{*}}\phi_{i^{*}}y_{i^{*}}.
		\end{equation} 
		Furthermore, we can observe that the marginal gain of the greedy search is then proportional to $\lambda_{i^{*}}\|\phi_{i^{*}}\|$ of the candidate label $i^{*}$. To minimize the marginal gain, we choose the candidate label $i^{*}$ producing the minimal $\lambda_{i^{*}}\|\phi_{i^{*}}\|$.
	\end{observation}
	
	By taking $\frac{\partial{J}}{\partial{\lambda_{i}}} = 0$ and substituting the above expression of $r$, we can obtain:
	\begin{equation}\label{eq:lambdar}
	\lambda_{i^{*}} = \frac{2(t_{i^{*}} + y_{i^{*}}h_{i^{*}}(\tilde{x}))}{\|\phi_{i^{*}}\|^2_{2}}.
	\end{equation}
	Consequently, we can derive that the upper bound of the required adversarial perturbation norm $r_{-i^{*}}$ as follows: 
	\begin{equation}\label{eq:rbound}
	\frac{|y_{i^{*}}h_{i^{*}}(\tilde{x}) + t_{i^{*}}|}{\|\phi_{i^{*}}\|_{2}}\geq\|r_{-i^{*}}\|_{2}. 
	\end{equation}
	As indicated by Eq.\ref{eq:rbound} and $t_{i} >0$, $\|r_{-i^{*}}\|_{2}$ is proportional to the ratio $\frac{|y_{i^{*}}h_{i^{*}}(\tilde{x})|}{\|\phi_{i^{*}}\|_{2}}$. It gives the conclusion of Lemma.2 in Section.4. Based on Eq.\ref{eq:optimalr} and Eq.\ref{eq:lambdar}, we can find that $\|\frac{\partial{\|r_{-i^{*}}\|_{2}}}{y_{i}}\|_{2}={\frac{|y_{i}h_{i}(\tilde{x}) + t_{i}|}{\|\phi_{i}\|_{2}}}$. Therefore, the greedy feed-forward expansion can be considered as conducting orthogonal matching pursuit based greedy search for the submodular function maximization problem \cite{ElenbergAA2016}. 
	
	
	\section{Experimental Setup}
	\subsection{Dataset Information}
	We include 4 datasets collected from various real-world multi-label cyber security practices (\textit{Creepware}) biology research (\textit{Genbase})\cite{Tsoumakas2010}, object recognition (\textit{VOC2012})\cite{pascal-voc-2012} and environment research (\textit{Planet}\cite{planet2017}). Except from the well-known \textit{VOC2012} dataset, \textit{Creepware} data include 966 stalkware app instances intercepted by the mobile AV service of a private security vendor. Each app has 16 labels indicating different types of surveillance on the victim's mobile device. The surveillance types include malicious remote control functions, such as recording messages/phone calls/call logs, logging key pressing, tracking GPS locations, extracting photos, remotely accessing cameras of the victim's mobile device and so on. Each app is profiled by the introductory texts of the app available in the third-party app stores and signatures of its mobile service access. \textit{GenBase} dataset contains 662 proteins. Each protein molecule may belong to one or more classes among the 10 protein families concerned in a bio-medicine clinical study. One protein molecule is described by a binary string, denoting whether or not a specific signature of the molecule structure is present. \textit{Planet} data collects daily satellite imagery of the entire land surface of the earth at 3-5 meter resolution. Each image is equipped with labels denoting different atmospheric conditions and various classes of land cover/land use.
	
	\subsection{Implementation Platforms}
	\textbf{Software Platform}: Our codes were implemented in Python and all the models were built by Keras package. \textbf{Hardware Platform:} Our experiments were conducted on GPU rtx2080ti.
	
	\subsection{Targeted Classifiers}
	To instantiate the attackability analysis, we study empirically two types of the targeted multi-label classifiers: linear Support Vector Machine (SVM) and Deep Neural Nets (DNN) based classifier. On \textit{Creepware}, linear SVM is applied on the TF-IDF feature vectors extracted from the descriptive texts and categorical service access signatures. On \textit{Genbase}, we use directly the binary strings as input features for classification. The DNN models of Inception-V3 based and Resnet50 based multi-label classifiers are used for \textit{VOC2012} and \textit{Planet}. Specially, we replace the last layer of Inception-V3 and Resnet50 by $m$ logistic regression structures to build multi-label classifiers. 
	We show the performance of Resnet50 classifiers on unperturbed test instances in Table \ref{tab:resnet}.
	
	\begin{table}[!h]
		\small                    
		\setlength\tabcolsep{2.0pt} 
		\caption {
			The F1-scores of Resnet50 based classifiers on different datasets}	\small 
		\label{tab:resnet}
		\newcommand{\tabincell}[2]{\begin{tabular}{@{}#1@{}}#2\end{tabular}}
		\centering
		\begin{tabular}{c|c|c|c|c|c|c} 
			\hline
			\hline
			Dataset	& $N$ & m & ${l}_{avg}$ & Micro F1 & Macro F1 & Classifier$_{target}$\\
			\hline
			{VOC2012}	&17,125 	&20 & 1.39  & 0.79 &0.68  & Resnet50  \\
			\hline
			{Planet}	&40,479 	&17 & 2.87   & 0.78  &0.35   & Resnet50   \\
			\hline
			\hline
		\end{tabular}
	\end{table}

	
	\subsection{Attack and Adversarial Training}
	In addition to \textit{Projected gradient decent} (PGD) \cite{Madry2018iclr} based attack method, we also adopt \textit{Carlini-Wagner} attack (CW) to conduct the step of targeted attack of Algorithm.\ref{alg:estimator} in our study on the dataset \emph{Creepware}.
	
	
	
	Our theoretical and empirical evaluation of attackability are both defined regardless of the concrete choices of evasion attack method. Correspondingly, we involve different attack methods at the core of the label space exploration. The aim is to verify the attack-method-independence of the proposed attackability analysis.

	\subsection{Performance Benchmark}
	We vary the limit of the attack budget $\varepsilon$ to denote the attack strength. Given a fixed value of $\varepsilon$, we calculate \textbf{the average number of flipped labels on test data as an estimator of the empirical classification risk $R^{emp}_{\mathcal{P}'}$ induced by the attack. This is defined as the empirical attackability indicator, as given in the design of fast greedy attack space exploration}. 
	
	\textbf{First}, to evaluate the effectiveness of the proposed empirical attackability indicator, we compute the averaged number of flipped labels only on the test instances of which all the labels are correctly classified. The propose is to assess whether the proposed greedy expansion method can identify significantly more labels as the feasible attack target, compared to well established baseline exploration strategies. \textbf{Second}, in order to verify the theoretical analysis over the countermeasures in the attackability assessment, we calculate the averaged number of miss-classified labels caused by the adversarial perturbation over all the testing data instances. It is consistent with the definition of the empirical attackability estimator in Eq.\ref{emrisk}. The resultant empirical attackability indicator is used to demonstrate whether the countermeasures can help to mitigate the threat of evasion attack. 
	
	\subsection{Baselines in Validation of Empirical Attackability Indicator}
	We assess the effectiveness of the proposed greedy expansion based empirical attackability indicator. Especially, we involve four baselines of label exploration strategies to search for maximal label perturbation. 
	\begin{itemize}
		\item \textbf{PGS} (Primitive Greedy Search): In each round of the greedy search, the primitive greedy search method calculates the magnitude of r with the combination of the current set $S$ and each of the candidate labels. Then it chooses the label contributing the least increasing of $\|r\|^2$. Though PSG can achieve the exact greedy search, it requires to run evasion attack for each candidate label. Therefore it is significantly heavier than the proposed \textbf{GASE} method.  
		\item \textbf{RS} (Random Search): In each round of RS, one label is selected purely randomly from the candidate set and added to the current set $S$. Randomized search doesn't pursue to optimize the exploration objective function in Eq.\ref{eq:bilevelopt}. It is involved to show the necessity of the heuristic rule in the label exploration, such as the principle of the greedy search.     
		\item \textbf{OS} (Oblivious Search): The oblivious method doesn't conduct iterative expansion of the set. This method first compute the norm of the adversarial perturbation induced by flipping each candidate label and keeping the other labels unchanged. The labels causing the least perturbation magnitudes are selected to form the set $S$. It is required to check if flipping all the labels in $S$ can deliver a feasible evasion attack. 
		\item \textbf{LS} (Loss-guided Search) : In each iteration of the LS method, it updates the adversarial perturbation $r$ along the direction where the multi-label classification loss is increased the most. The iterative update of $r$ is stopped until $\|r\|^2$ surpasses the cost limit. The set of the attacked labels given the derived $r$ are reported. \textbf{LS} doesn't use any attack method in its implementation. It is simply a gradient ascent process. Maximizing the loss only, though sounding reasonable, is a rough search strategy. The increasing of the loss can be caused by pushing originally miss-classified labels even further from the correct decision, instead of flipping originally correctly predicted labels. As a result, it misleads the search of the attackable labels. 
	\end{itemize}

	\subsection{The Influence of Imposed Nuclear Norm Constrain on Clean Test Data}
	We adopt the tech in \cite{nuclearnorm2019} to impose nuclear norm constraint. Specially, its implementation in package \emph{tensorflow-addons} was adopted in our experiments.
	Our experimental observation in Table.\ref{claclean} shows that  there is a obvious trade-off between improving a classifier's robustness by imposing the nuclear norm constraint and preserving its good utility on unperturbed test data. A strong nuclear norm constraint improves greatly the adversarial robustness. Nevertheless, it also causes accuracy loss of the classifier.
	
	\begin{table*}[!htb]
		\small
		\setlength\tabcolsep{0pt}
		\caption {Classification results on adversary-free test data. The classifiers are trained with increasingly stronger nuclear norm based constraints}
		\label{claclean}
		\renewcommand\arraystretch{1.0}
		\resizebox{1.0\textwidth}{!}{ 
			\newcommand{\tabincell}[2]{\begin{tabular}{@{}#1@{}}#2\end{tabular}}
			\centering
			\begin{tabular}{|c||*{6}{c|}}\hline
				\backslashbox{Dataset}{Classifier}
				&No constraint&$\lambda = 1e-4$&{$\lambda = 1e-3$}&{$\lambda = 5e-2$}
				&{$\lambda = 1e-2$}&{$\lambda = 1e-1$}\\\hline\hline
				\textit{Creepware SVM} &$\begin{array}{l}
				micro\_{F_1}:0.760\\
				macro\_{F_1}:0.662
				\end{array}$&$\begin{array}{l}
				micro\_{F_1}:0.764\\
				macro\_{F_1}:0.646
				\end{array}$&$\begin{array}{l}
				micro\_{F_1}:0.752\\
				macro\_{F_1}:0.620
				\end{array}$&$\begin{array}{l}
				micro\_{F_1}:0.679\\
				macro\_{F_1}:0.476
				\end{array}$&$\begin{array}{l}
				micro\_{F_1}:0.600\\
				macro\_{F_1}:0.343
				\end{array}$&$\begin{array}{l}
				micro\_{F_1}:0.471\\
				macro\_{F_1}:0.110
				\end{array}$\\\hline
				\textit{Genbase SVM} &$\begin{array}{l}
				micro\_{F_1}:0.991\\
				macro\_{F_1}:0.733
				\end{array}$&$\begin{array}{l}
				micro\_{F_1}:0.994\\
				macro\_{F_1}:0.737
				\end{array}$&$\begin{array}{l}
				micro\_{F_1}:0.991\\
				macro\_{F_1}:0.725
				\end{array}$&$\begin{array}{l}
				micro\_{F_1}:0.982\\
				macro\_{F_1}:0.677
				\end{array}$&$\begin{array}{l}
				micro\_{F_1}:0.929\\
				macro\_{F_1}:0.546
				\end{array}$&$\begin{array}{l}
				micro\_{F_1}:0.634\\
				macro\_{F_1}:0.209
				\end{array}$\\\hline
				\textit{VOC2012 Inception-V3} &$\begin{array}{l}
				micro\_{F_1}:0.827\\
				macro\_{F_1}:0.736
				\end{array}$&$\begin{array}{l}
				micro\_{F_1}:0.826\\
				macro\_{F_1}:0.727
				\end{array}$&$\begin{array}{l}
				micro\_{F_1}:0.825\\
				macro\_{F_1}:0.736
				\end{array}$&$\begin{array}{l}
				micro\_{F_1}:0.823\\
				macro\_{F_1}:0.725
				\end{array}$&$\begin{array}{l}
				micro\_{F_1}:0.819\\
				macro\_{F_1}:0.709
				\end{array}$&$\begin{array}{l}
				micro\_{F_1}:0.602\\
				macro\_{F_1}:0.143
				\end{array}$\\\hline
				\textit{Planet Inception-V3} &$\begin{array}{l}
				micro\_{F_1}:0.822\\
				macro\_{F_1}:0.361
				\end{array}$&$\begin{array}{l}
				micro\_{F_1}:0.818\\
				macro\_{F_1}:0.354
				\end{array}$&$\begin{array}{l}
				micro\_{F_1}:0.823\\
				macro\_{F_1}:0.360
				\end{array}$&$\begin{array}{l}
				micro\_{F_1}:0.819\\
				macro\_{F_1}:0.355
				\end{array}$&$\begin{array}{l}
				micro\_{F_1}:0.819\\
				macro\_{F_1}:0.347
				\end{array}$&$\begin{array}{l}
				micro\_{F_1}:0.695\\
				macro\_{F_1}:0.192
				\end{array}$\\\hline
				\textit{VOC2012 Resnet50} &$\begin{array}{l}
				micro\_{F_1}:0.788\\
				macro\_{F_1}:0.683
				\end{array}$&$\begin{array}{l}
				micro\_{F_1}:0.783\\
				macro\_{F_1}:0.670
				\end{array}$&$\begin{array}{l}
				micro\_{F_1}:0.785\\
				macro\_{F_1}:0.677
				\end{array}$&$\begin{array}{l}
				micro\_{F_1}:0.780\\
				macro\_{F_1}:0.672
				\end{array}$&$\begin{array}{l}
				micro\_{F_1}:0.773\\
				macro\_{F_1}:0.656
				\end{array}$&$\begin{array}{l}
				micro\_{F_1}:0.643\\
				macro\_{F_1}:0.261
				\end{array}$\\\hline
				\textit{Planet Resnet50} &$\begin{array}{l}
				micro\_{F_1}:0.778\\
				macro\_{F_1}:0.352
				\end{array}$&$\begin{array}{l}
				micro\_{F_1}:0.790\\
				macro\_{F_1}:0.355
				\end{array}$&$\begin{array}{l}
				micro\_{F_1}:0.794\\
				macro\_{F_1}:0.353
				\end{array}$&$\begin{array}{l}
				micro\_{F_1}:0.804\\
				macro\_{F_1}:0.357
				\end{array}$&$\begin{array}{l}
				micro\_{F_1}:0.795\\
				macro\_{F_1}:0.335
				\end{array}$&$\begin{array}{l}
				micro\_{F_1}:0.710\\
				macro\_{F_1}:0.219
				\end{array}$\\\hline
			\end{tabular}
		}
	\end{table*}
	
	\subsection{Validation of Empirical Attackability Indicator on Resnet50}
\begin{figure}[!htb]
		\centerline{\includegraphics[width=0.9\columnwidth]{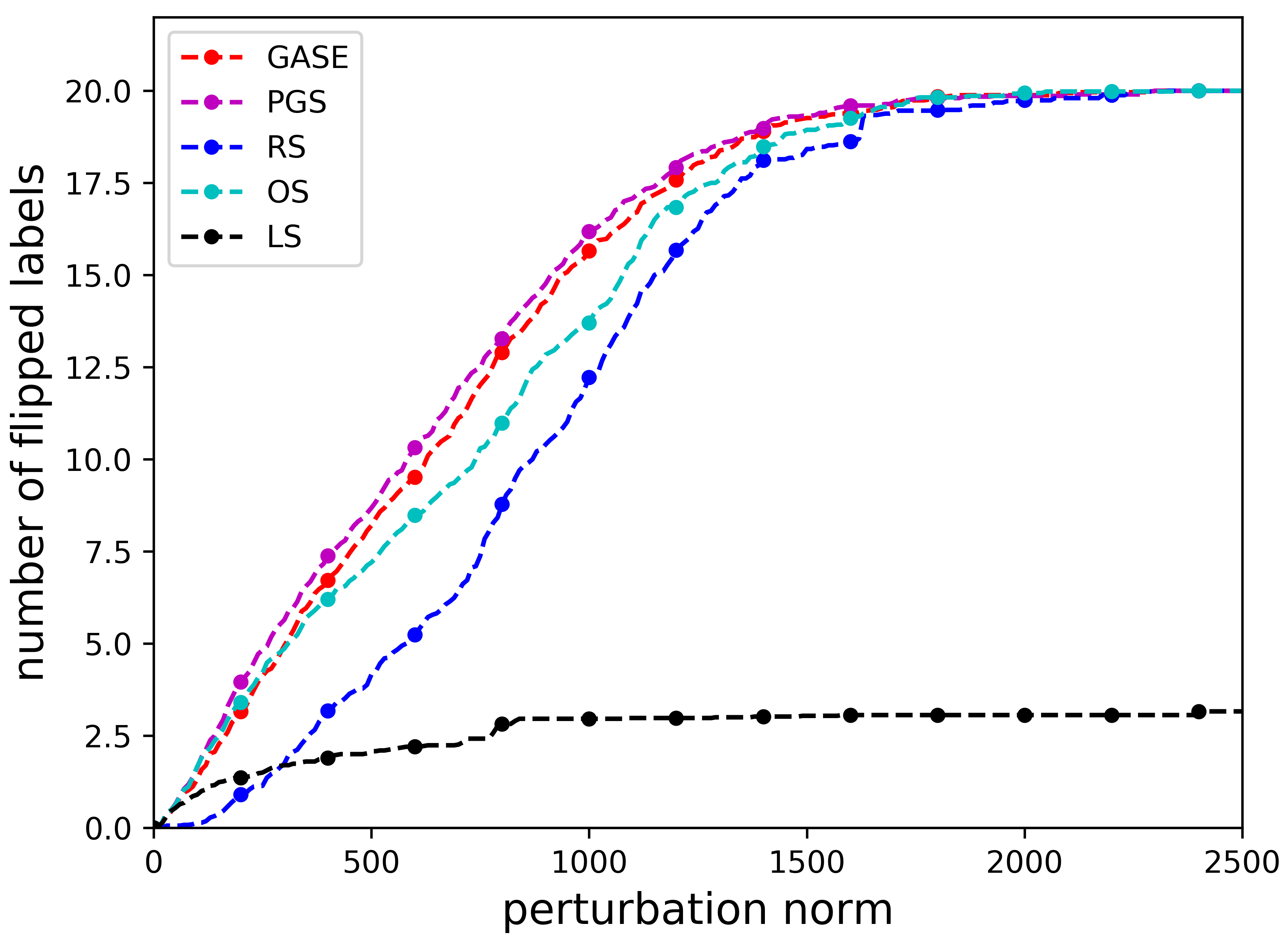}}
		\caption{The empirical attackability indicator estimated by different label exploration strategies. Attackability of Resnet50 on \emph{VOC2012}}
		\label{rankresnetvoc}
	\end{figure}

\begin{figure}[!htb]
		\centerline{\includegraphics[width=0.9\columnwidth]{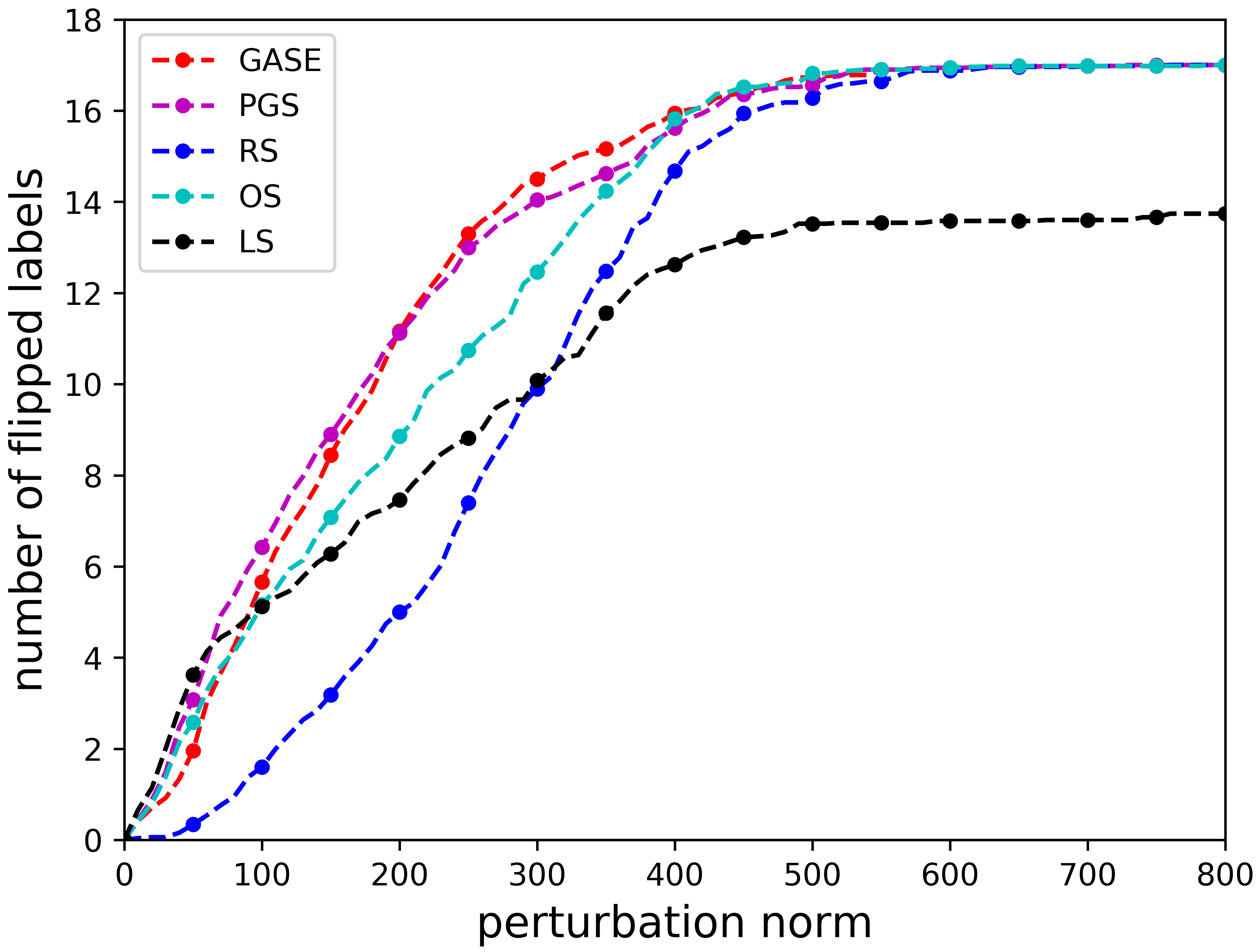}}
		\caption{The empirical attackability indicator estimated by different label exploration strategies. Attackability of Resnet50 on \emph{Planet}}
		\label{rankresnetplt}
	\end{figure}

	Fig.\ref{rankresnetvoc} and \ref{rankresnetplt} show the number of flipped labels obtained by the proposed \textit{GASE} algorithm and the baselines on Resnet50 based deep multi-label classifiers over dataset \emph{VOC2012} and \emph{Planet}. The results verify again the reasonableness of our greedy search strategy stated in Lemma 1, since GASE and PGS achieve mre flipped labels than other baselines. 
	
	\subsection{Attackability Evaluation with Countermeasures for Evasion Attack on Resnet50}
\begin{figure}[!htb]
		\centerline{\includegraphics[width=0.9\columnwidth]{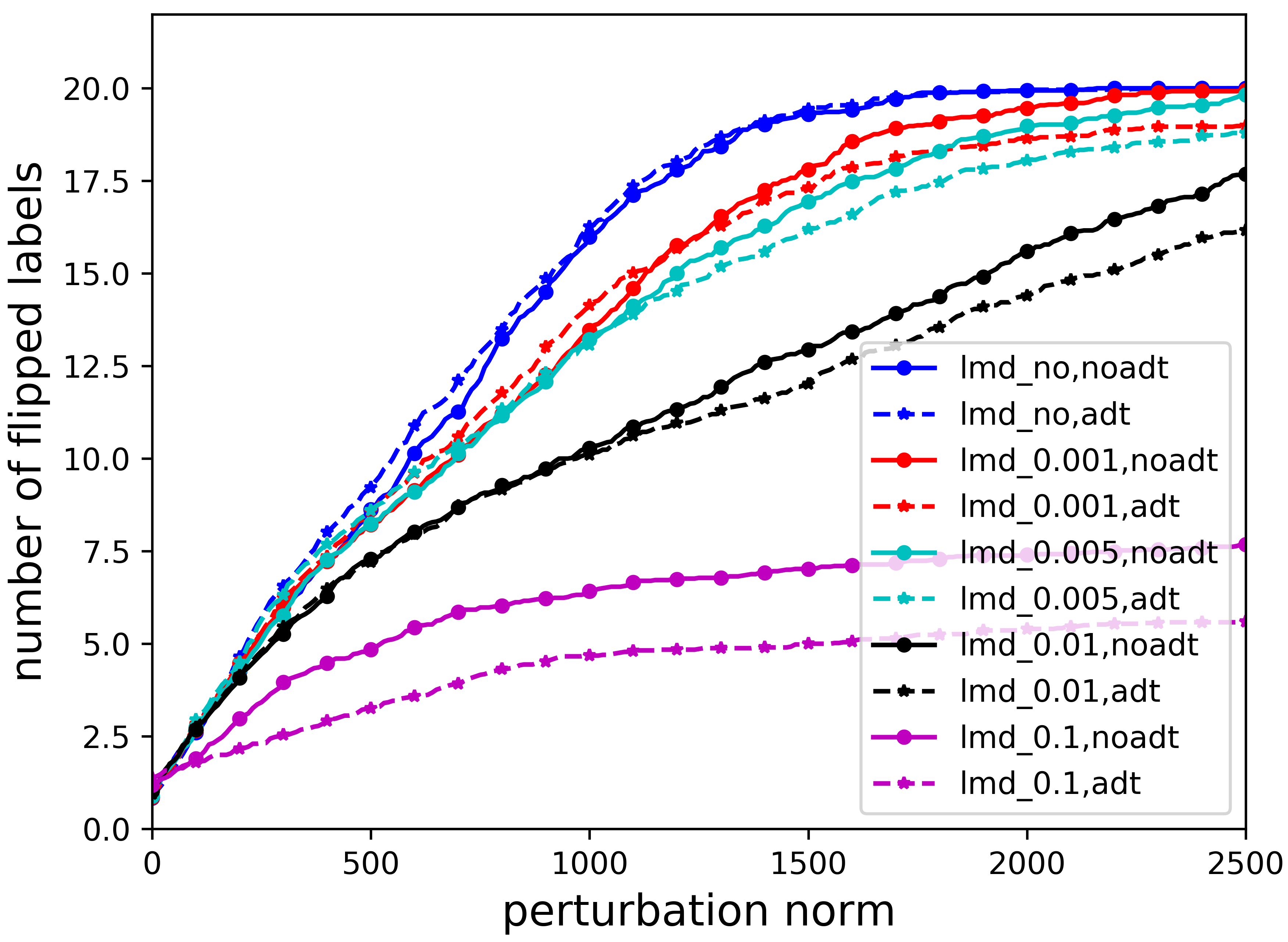}}
		\caption{The  evaluation of classifiers' attackability under different complexity controls.. Attackability of controlled Resnet50 on \emph{VOC2012}}
		\label{modelresnetvoc}
	\end{figure}

\begin{figure}[!htb]
		\centerline{\includegraphics[width=0.9\columnwidth]{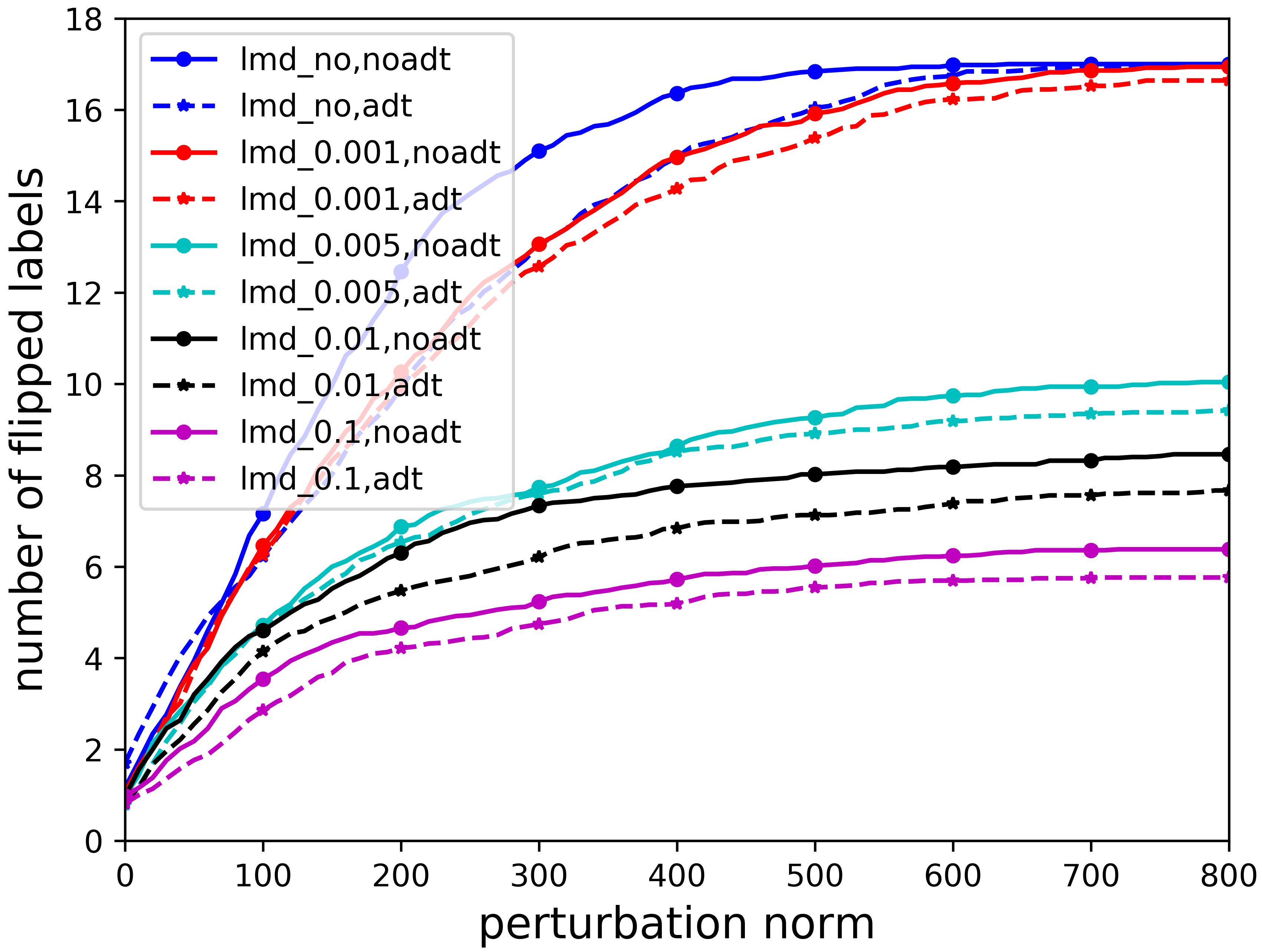}}
		\caption{The  evaluation of classifiers' attackability under different complexity controls.. Attackability of controlled Resnet50 on \emph{Planet}}
		\label{modelresnetplt}
	\end{figure}

	Fig.\ref{modelresnetvoc} and \ref{modelresnetplt} show the attackability evaluation results of Resnet based classifiers on dataset \emph{VOC2012} and \emph{Planet} under different complexity controls and adversarial training.  The trend is similar to the results of Inception-V3 based classifiers, that is in general, reducing model complexity and adversarial training can improve classifiers' adversarial robustness.  
	
	\subsection{Case Study on Dataset \emph{Creepware} with CW Attack}
\begin{figure}[!htb]
		\centerline{\includegraphics[width=0.9\columnwidth]{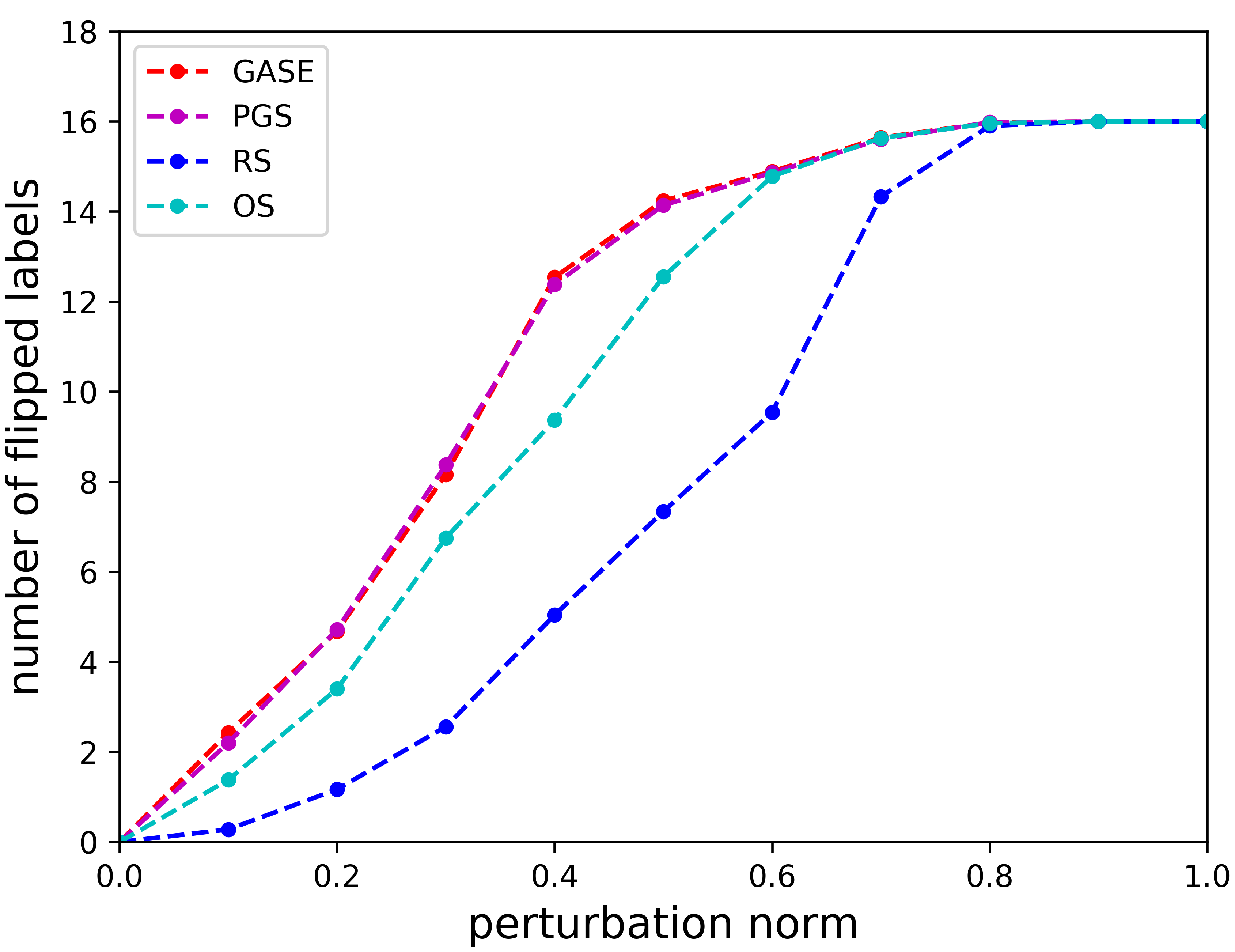}}
		\caption{The empirical attackability indicator estimated by different label exploration strategies. Attackability of SVM on \emph{Creepware}. The targeted evasion attack is achieved by CW attack.}
		\label{rankcwcreep}
	\end{figure}

\begin{figure}[!htb]
		\centerline{\includegraphics[width=0.9\columnwidth]{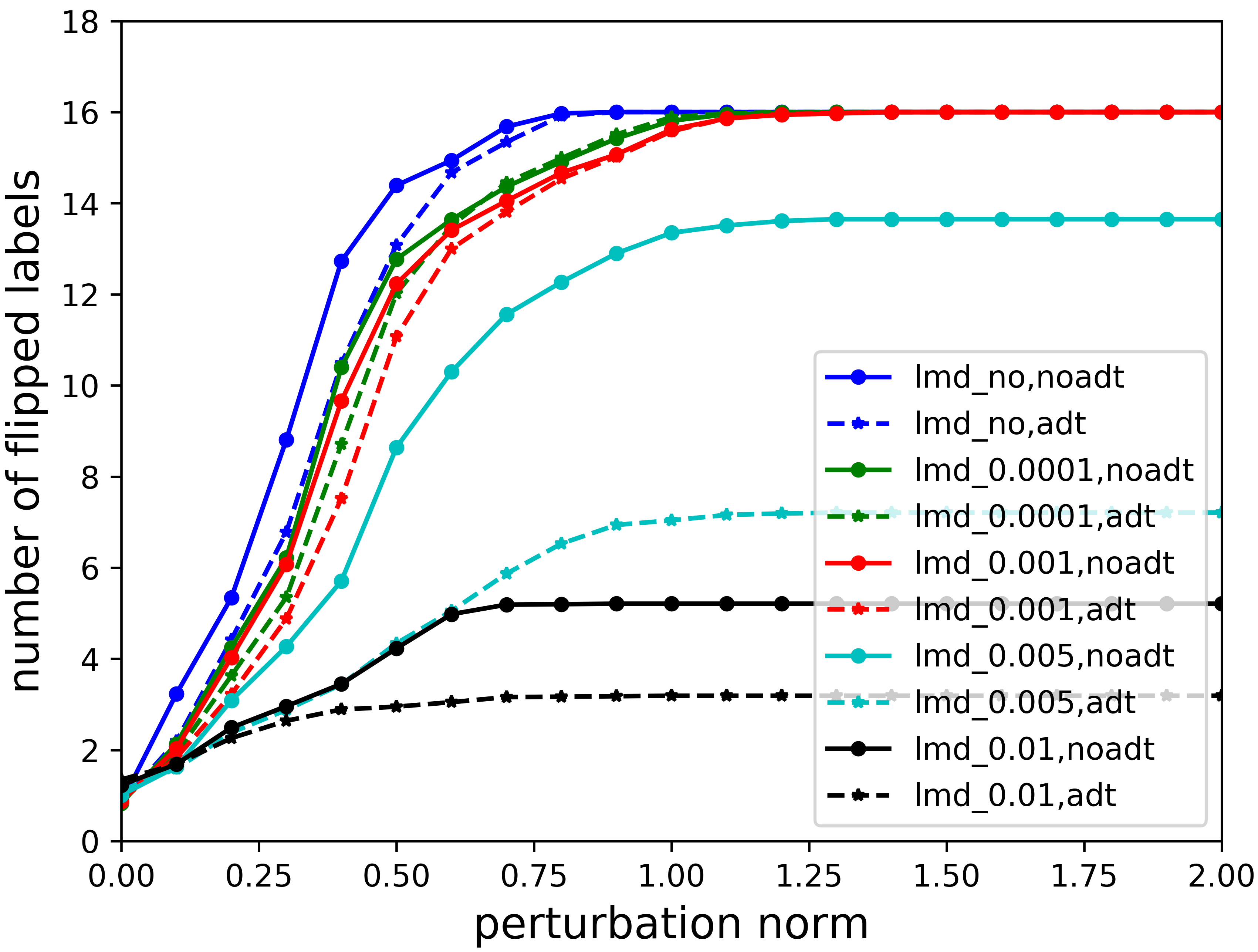}}
		\caption{The  evaluation of classifiers' attackability under different complexity controls. Attackability of controlled SVM on \emph{Creepware}. The targeted evasion attack is achieved by CW attack.}
		\label{modelcwcreep}
	\end{figure}

	We replace the targeted evasion attack method PGD used in previous experiments by Carlini-Wagner (CW) to verify the attack-method-independence of 1) our proposed greedy based label exploration strategy and 2) the attackability analysis in terms of controlling model complexity and adversarial training. Fig. \ref{rankcwcreep} and \ref{modelcwcreep} show the indicator estimation and attackability evaluation results of SVM based classifiers on dataset \emph{Creepware} respectively, notably, here we used CW attack to achieve the targeted evasion attack. The results are similar to the results with PGD attack, especially the indicator estimation results. In Fig. \ref{rankcwcreep}, the result of LS baseline is not included, as this baseline is independent of targeted evasion attack method.  
	
	In Fig. \ref{modelcwcreep}, when the nuclear norm-based constraint is strong and adversarial training is conducted simultaneously, the number of flipped labels using the proposed attackability indicator with CW attack is smaller than that derived by using the PGD-based attack (showed in Fig. \ref{permodel-all}). The reason is that we constraint the range of line search tuning in CW attack to control the time cost of the attack step.
	
	Notably, the results in Fig. \ref{modelcwcreep} confirm the improvement of adversarial robustness by introducing adversarial training and the low-rank constraint over the classifier's parameters. These two approaches reduces the empirical worst-case risk over the adversarial samples and controls the classifier's complexity respectively, which in turn reduces the expected worst-case risk. The results are consistent with our theoretical analysis of the classifier's attackability. Furthermore, we confirm that the empirical attackability evaluation can use any targeted evasion attack method as its component. 
	
\end{document}